\documentclass[journal,transmag]{IEEEtran}

\usepackage{graphicx}
\usepackage{amsmath}
\usepackage{amssymb}
\usepackage{booktabs}
\usepackage{multirow}
\usepackage{subcaption}
\usepackage{tikz}
\usepackage{float}
\usepackage{comment}

\usepackage[final]{pdfpages}

\usepackage[pagebackref,breaklinks,colorlinks]{hyperref}

\usepackage[capitalize]{cleveref}
\crefname{section}{Sec.}{Secs.}
\Crefname{section}{Section}{Sections}
\Crefname{table}{Table}{Tables}
\crefname{table}{Tab.}{Tabs.}

\begin{document}

\title{CR-FIQA: Face Image Quality Assessment by Learning Sample Relative Classifiability}

\author{Fadi Boutros$^{1}$, Meiling Fang$^{1,2}$,Marcel Klemt$^{1}$, Biying Fu$^{1}$, Naser Damer$^{1,2}$\\
$^{1}$Fraunhofer Institute for Computer Graphics Research IGD, Darmstadt, Germany\\
$^{2}$Department of Computer Science, TU Darmstadt,
Darmstadt, Germany\\
Email: fadi.boutros@igd.fraunhofer.de
}

\maketitle

\begin{abstract}
Face image quality assessment (FIQA) estimates the utility of the captured image in achieving reliable and accurate recognition performance.
This work proposes a novel FIQA method, CR-FIQA, that estimates the face image quality of a sample by learning to predict its relative classifiability.
This classifiability is measured based on the allocation of the training sample feature representation in angular space with respect to its class center and the nearest negative class center.
We experimentally illustrate the correlation between the face image quality
and the sample relative classifiability.
As such property is only observable for the training dataset, we propose to learn this property by probing internal network observations during the training process and utilizing it to predict the quality of unseen samples.
Through extensive evaluation experiments on eight benchmarks and four face recognition models, we demonstrate the superiority of our proposed CR-FIQA over state-of-the-art (SOTA) FIQA algorithms.
\footnote{\url{https://github.com/fdbtrs/CR-FIQA}}
\end{abstract}

\section{Introduction}
Face image utility indicates the utility (value) of an image to face recognition (FR) algorithms \cite{ISOIEC29794-1,best2018learning}. 
This utility is measured with a scalar, namely the face image quality (FIQ) score, following the definition in ISO/IEC 2382-37 \cite{ISOIEC2382-37} and the FR Vendor Test (FRVT) for FIQA \cite{NISTQuaity}.

As FIQA measures the face utility to FR algorithm, it does not necessary reflects, and does not aim at measuring, the perceived image quality, e.g. a profile face image can be of high perceived quality but of low utility to FR algorithm \cite{DBLP:conf/icb/TerhorstKDKK20}.
Assessing this perceived image quality has been addressed in the literature by general image quality assessment (IQA) methods \cite{BRISQE_IQA,nique,liu2017rankiqa} and is different than assessing the utility of an the image for FR .
This is reflected by FIQA methods \cite{MagFace,SDDFIQA,SERFIQ} significantly outperforming IQA methods \cite{BRISQE_IQA,nique,liu2017rankiqa} in measuring the utility \cite{ISOIEC29794-1} of face images in FR, as demonstrated in \cite{MagFace,SERFIQ,BiyingWACV}. 



SOTA FIQA methods focused either on creating concepts to label the training data with FIQ scores and then learn a regression problem \cite{SDDFIQA,hernandez2019faceqnet,faceqnetv1}, or on developing a link between face embedding properties under certain scenarios and the FIQ \cite{SERFIQ,MagFace,PFE_FIQA}.
Generally, the second approach led to better FIQA performances with most works mentioning the error-prone labeling of the ground truth quality in the first research direction as a possible reason \cite{SERFIQ,MagFace}.
However, in the second category, transferring the information in network embeddings into an FIQ score is not a learnable process, but rather a form of statistical analysis, which might not be optimal.

\begin{figure*}[ht!]
\begin{center}
\includegraphics[width=0.75\linewidth]{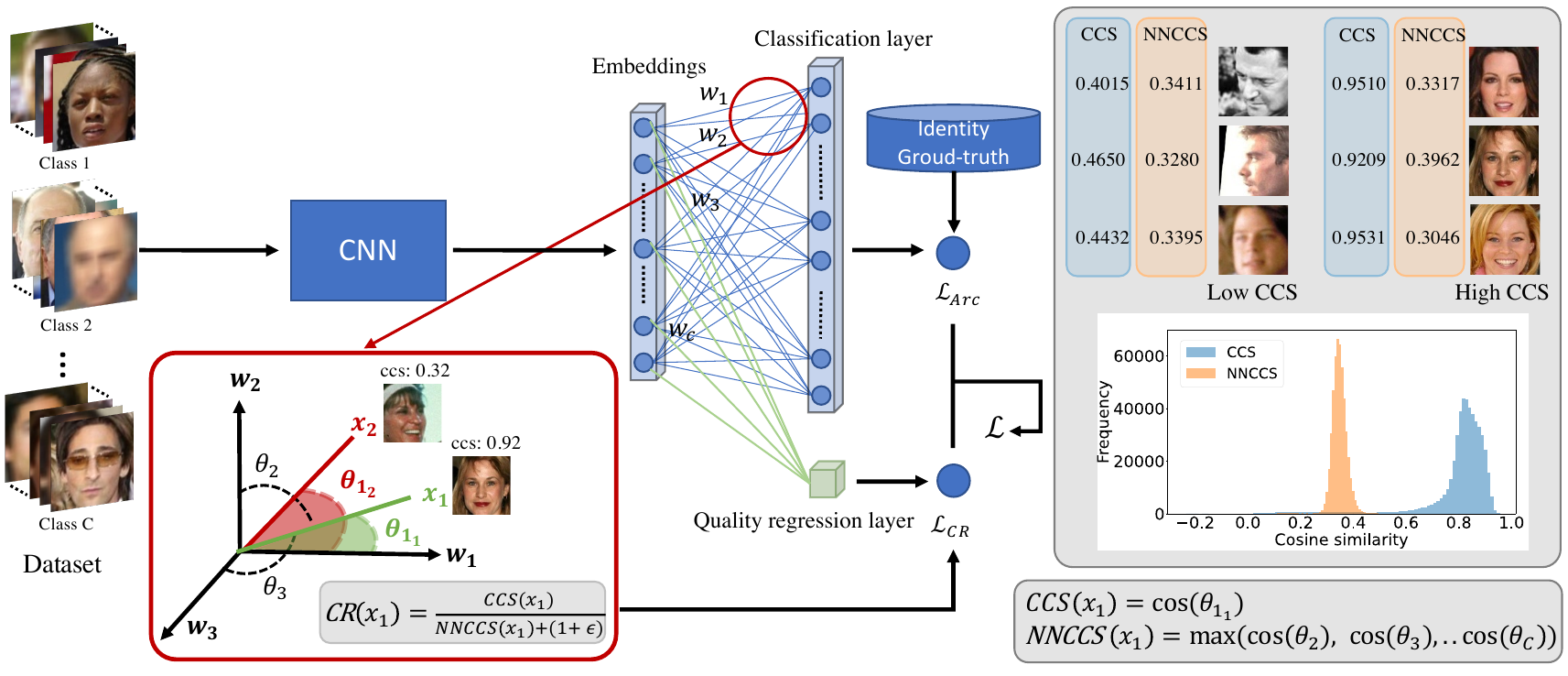}
\end{center}
\caption{An overview of our CR-FIQA training paradigm. 
We propose to simultaneously learn to optimize the class center ($\mathcal{L}_{Arc}$), while learning to predict an internal network observation i.e. the allocation of the feature representation of sample $x$ in feature space, with respect to its class center $w_1$ and nearest negative class center $w_2$ ($\mathcal{L_{CR}}$). 
The figure in the red rectangle illustrates the angle between two samples $x_1$ and $x_2$ (belong to identity 1) and their class center $w_1$.  
The plot on the right of the figure shows the distribution of the cosine similarity between training samples and their class centers (CCS) and nearest negative class centers (NNCCS) obtained from ResNet-50 trained on CASIA-WebFace \cite{casia_webface}. The example images on the top-right of this plot are of high CCS values and the ones on the top-left are of low CCS values (notice the correspondence to perceived quality). These samples are selected from CASIA-WebFace \cite{casia_webface}. During the testing mode, the classification layer is removed and the output of the regression layer is used to predict the FIQA of testing samples.  }
\label{fig:workflow}
\end{figure*}

This paper proposes a novel learning paradigm to assess FIQ, namely the CR-FIQA.
Our concept is based on learning to predict the classifiability of FR training samples by probing internal network observations that point to the relative proximity of these samples to their class centers and negative class centers.
This regression is learned simultaneously with a conventional FR training process that minimizes the distance between the training samples and their class centers.
Linking the properties that cause high/low classifiability of a training sample to the properties leading to high/low FIQ, we can use our CR-FIQA to predict the FIQ of any given sample. 
We empirically prove the theorized link between classifiability (Section \ref{sec:relation_CR_FIQA}) and FIQ and conduct thorough ablation studies on key aspects of our CR-FIQA design (Section \ref{sec:abl}).
The proposed CR-FIQA is evaluated on eight benchmarks along with SOTA FIQAs.  
The reported results on four FR models demonstrate the superiority of our proposed CR-FIQA over SOTA methods and the stability of its performance across different FR models. 
An overview of the proposed CR-FIQA is presented in Figure \ref{fig:workflow} and will be clarified in detail in this paper.

\section{Related work}




The recent SOTA FIQA approaches can be roughly grouped into two main categories. 
The first are approaches that learn a straight forward regressions problem to assess a FIQ score \cite{best2018learning,hernandez2019faceqnet,faceqnetv1,SDDFIQA,pcnet}. 
The second category uses properties of the FR model responses to face samples to estimate the sample quality without explicitly learning a typical supervised regression that requires quality labels \cite{SERFIQ,MagFace,PFE_FIQA}.
In the first category, the innovation focused on creating the FIQ labels for training. 
These quality labels included human-labeled quality labels \cite{best2018learning}, the FR genuine comparison score
between a sample and an ICAO \cite{ICAO18} compliant sample \cite{hernandez2019faceqnet,faceqnetv1}, the FR comparison score involving the labeled sample (assumed to have the lower quality in the comparison pair) \cite{pcnet}, and the Wasserstein distance between a randomly selected genuine and imposter FR comparisons with the labeled sample \cite{SDDFIQA}.
These solutions generally trained a regression network to predict the quality label, using both, trained-from-scratch networks in some cases \cite{pcnet}, and pre-trained FR networks in other cases \cite{SDDFIQA,hernandez2019faceqnet,faceqnetv1}.
A slightly different approach, however also based on learning from labels, focuses on learning to predict the sample quality as a rank \cite{learntorank} based on FR performance-based training rank labels of a set of databases \cite{RANKIQ_FIQA}.
In the second category, the innovation was rather focused on linking face embedding properties under certain scenarios to the FIQ, without the explicit need for quality-labeled data. 
In \cite{SERFIQ}, the assessed sample is passed through an FR network multiple times, each with a different random dropout pattern. The robustness of the resulting embeddings, represented by the sigmoid of the negative mean of the Euclidean distances between the embeddings, is considered the FIQ score.
In \cite{MagFace}, the FIQ score is calculated as the magnitude of the sample embedding. This is based on training the FR model using a loss that adapts the penalty margin loss based on this magnitude, and thus links the closeness of a sample to its class center to the unnormalized embedding magnitude.
While in \cite{PFE_FIQA}, the solution produces both, an FR embedding and a gaussian variance (uncertainty) vector, from a face sample. 
The inverse of harmonic mean of the uncertainty vector is considered as the FIQ score.
Our CR-FIQA learns a regression problem to estimate the FIQ score, however, unlike previous works, without relying on preset labels, but rather learn a dynamic internal network observations (during training) that point out sample classifiability.






\section{Approach} 
This section presents our proposed Certainty Ratio Face Image Quality Assessment (CR-FIQA) approach, which inspects internal network observations to learn to predict the sample relative classifiability. This classifiability prediction is then used to estimate the FIQ. 
An overview of the proposed CR-FIQA approach is presented in Figure \ref{fig:workflow}.
During the training phase of an FR model, the model can conveniently push the high-quality samples close to their class center and relatively far from other class centers.
Conversely, the FR is not able to push, to the same degree, low-quality samples to their class center, and thus they will remain relatively farther from their class center than the high-quality ones.
Based on this assumption, we theorize our approach by stating that the properties that cause a face sample to lay relatively closer to its class center during training are the ones that make it a high-quality sample, and vice versa. 
Therefore, learning to predict such properties in any given sample would lead to learning to assess this sample quality.
To learn to perform such assessment, our training paradigm targets learning internal network observations that evolve during the FR training phase, where these observations act as a training objective.
The predictions of such training paradigm can be simply stated as answering a question: if a given sample was hypothetically part of the FR model training (which it is not), how relatively close would it be to its class center? Answering this question would give us an indication of this sample quality as will be shown in detail in this paper. 

In the rest of this section, We formalize and empirically rationalize our proposed CR-FIQA approach and its components. To do that, we start by shortly revisiting angular margin penalty-based softmax loss utilized to optimize the class centers of the FR model. 
Then present a detailed description of our proposed CR-FIQA concept and the associated training paradigm.

\subsection{Revisiting Margin Penalty-based Softmax Loss}
Angular margin penalty-based softmax is a widely used loss function for training FR models \cite{deng2019arcface,curricularFace,MagFace,elasticface}. 
It extends over softmax loss by deploying angular penalty margin on the angle between the deep features and their corresponding weights. 
Margin penalty-based softmax loss aims to push the decision boundary of softmax, and thus enhance intra-class compactness and inter-class discrepancy. 
From this family of loss functions, this work utilizes ArcFace loss \cite{deng2019arcface} to optimize the distance between the training samples and their class center. 
Our choice of ArcFace loss is based on the SOTA performance achieved by ResNet-100 network trained with ArcFace on mainstream benchmarks \cite{deng2019arcface}.
Formally, ArcFace loss is defined as follows: 
\begin{equation}
\label{eq:arc}
    \mathcal{L}_{Arc}=\frac{1}{N}  \sum\limits_{i \in N} - log \frac{e^{s (cos(\theta_{y_i}+m))}}{ e^{s(cos(\theta_{y_i}+m))} +\sum\limits_{j=1 , j \ne y_i}^{C}  e^{s ( cos(\theta_{j}))}},
\end{equation}
where $N$ is the batch size, $C$ is the number of classes (identities), $y_i$ is the class label of sample $i$ (in range $[1,C]$), $\theta_{y_i}$ is the angle between the features $x_{i}$ and the $y_i$-th class center $w_{y_i}$. {$x_{i} \in R^d$}  is the deep feature embedding of the last fully connected layer of size $d$. $w_{y_i}$ is the $y_i$-th column of weights $W \in R^d_C$ of the classification layer.
$\theta_{y_i}$ is defined as ${x_iw^T_{y_i}}=\Vert x_i \Vert \Vert w_{y_i} \Vert cos(\theta_{y_i})$ \cite{DBLP:conf/cvpr/LiuWYLRS17}. The weights and the feature norms are fixed to $\Vert w_{y_i} \Vert=1$ and $ \Vert x_i \Vert =1$, respectively, using $l_2$ normalization as defined in \cite{DBLP:conf/cvpr/LiuWYLRS17,DBLP:conf/cvpr/WangWZJGZL018}. The decision boundary, in this case, depends on the angle cosine  between $x_{i}$ and $w_{y_i}$.
$m>0$ is an additive angular margin proposed by ArcFace \cite{deng2019arcface} to
enhance the intra-class compactness and inter-class discrepancy. Lastly, $s$ is the scaling parameter \cite{DBLP:conf/cvpr/WangWZJGZL018}.

\subsection{Certainty Ratio}
In this section, we formulate and empirically rationalize the main concepts that build our FIQA solution. 
We derive our Certainty Ratio (CR) to estimate the sample relative classifiability. 
Additionally, we experimentally illustrate the strong relationship between our CR measure and FIQ. 

\textbf{Certainty Ratio}
During the FR model training phase, the model is trained to enhance the separability between the classes (identities) by pushing each sample $x_{i}$ to be close to its class center $w_{y_i}$ and far from the other (negative) class centers $w_{j}, j \neq y_i$. 
Based on this, we first define the Class Center Angular Similarity (CCS) as the proximity between $x_{i}$ and its class center $w_{y_i}$, as follows:
\begin{equation}
CCS_{x_{i}}= cos(\theta_{y_i}),
\end{equation}
where $\theta_{y_i}$ is the angle between $x_{i}$ and its class center $w_{y_i}$, where the weights of the last fully connected layer of the FR model trained with softmax loss are considered as the centers for each class \cite{liu2017sphereface,deng2019arcface}.
Then, we define the Closest Nearest Negative Class Center Angular Similarity (NNCCS) as proximity between $x_{i}$ and the nearest negative class center $w_{j}, j \neq y_i$. 
Formally, NNCCS is defined as follows:
\begin{equation}
NNCCS_{x_{i}}= \max_{j=1 , j \ne y_i}^{C}(cos(\theta_{j})),
\end{equation}
where $\theta_{j}$ is the angle between $x_{i}$ and $w_{j}$.
As we theorize, when the FR model converges, the high-quality samples are pushed closer to their class centers (high CCS) 
in relation to their distance to neighbouring negative class centers (low NNCCS). 
However, low-quality samples can not be pushed as close to their class centers. 
A sample able to achieve high CCS with respect to NNCCS is a sample easily correctly classified during training, and thus is relatively highly classifiable.
We thus measure this relative classifiability by the ratio of CCS to NNCS, which we note as the Certainty Ratio (CR), as follows:
\begin{equation}
CR_{x_{i}}= \frac{CCS_{x_{i}}}{NNCCS_{x_{i}} +(1 + \epsilon)}, 
\end{equation}
where the $1+\epsilon$ term is added to insure a positive above zero denominator, i.e.  shift the NNCCS value range from $[-1,+1]$ to $[\epsilon,2+ \epsilon]$.
This ensures that the CR of a sample with a lower NNCCS is relatively higher than a sample with a higher NNCCS, given the same CCS, i.e. NNCCS regulates the CCS value in relation to neighbouring classes.
The $\epsilon$ is set to $1e-9$ in our experiments. 
The optimal CR is obtained when the CCS is approaching the maximum cosine similarity value (+1) and the NNCCS is approaching the minimum cosine similarity value (-1), i.e. the training sample is capable of being pushed to its class center, and far away from the closest negative class center, and thus it is highly classifiable.


\subsection{Relation between the CR and FIQ}
\label{sec:relation_CR_FIQA}
Here, we empirically prove the theorized relationship between the CR and FIQ (defined earlier as image utility).
Namely, we want to answer: if the CR values achieved by training samples of an FR model were used as FIQ, would they behave as expected from an optimal FIQ? If yes, then the face image properties leading to high/low CR do also theoretically lead to high/low FIQ.
To answer this question we conducted an experiment on a ResNet-50 \cite{DBLP:conf/cvpr/HeZRS16} FR model trained on CASIA-WebFace \cite{casia_webface} with ArcFace loss \cite{deng2019arcface} (noted as R50(CASIA)). 
Specifically, we calculate the CR, CCS, and NNCCS values from the trained model for all samples in the training dataset (0.5M images of 10K identities).
An insight on the resulting CCS and NNCCS values (CR being a derivative measure) is given as value distributions in Figure \ref{fig:workflow}, showing that these measures vary between different samples.
Furthermore, based on the calculated scores,  we plot Error vs.~Reject Curves (ERC) (described in Section \ref{sec:exp}) to demonstrate the relationship between the CR, as an FIQ measure, and FR performance.
To calculate the FR performance in the ERC curve, we extract the feature embedding of CASIA-WebFace \cite{casia_webface} using a ResNet-100 model \cite{DBLP:conf/cvpr/HeZRS16} trained on MS1M-V2 \cite{guo2016ms,deng2019arcface} with ArcFace (noted as R100(MS1M-V2)). 
We utilize a different model (trained on a different database) to extract the embedding (R100(MS1M-V2)) than the one used to calculate CR, CCS, and NNCCS (R50(CASIA)) to provide fair evaluation where the FR performance is evaluated on unseen data. 
Then, we perform $n:n$ comparisons between all samples of CASIA-WebFace using feature embedding obtained from the R100(MS1M-V2).

\begin{figure}[ht!]
  \centering
  \begin{subfigure}[b]{0.45\linewidth}
         \centering
   \includegraphics[width=\linewidth]{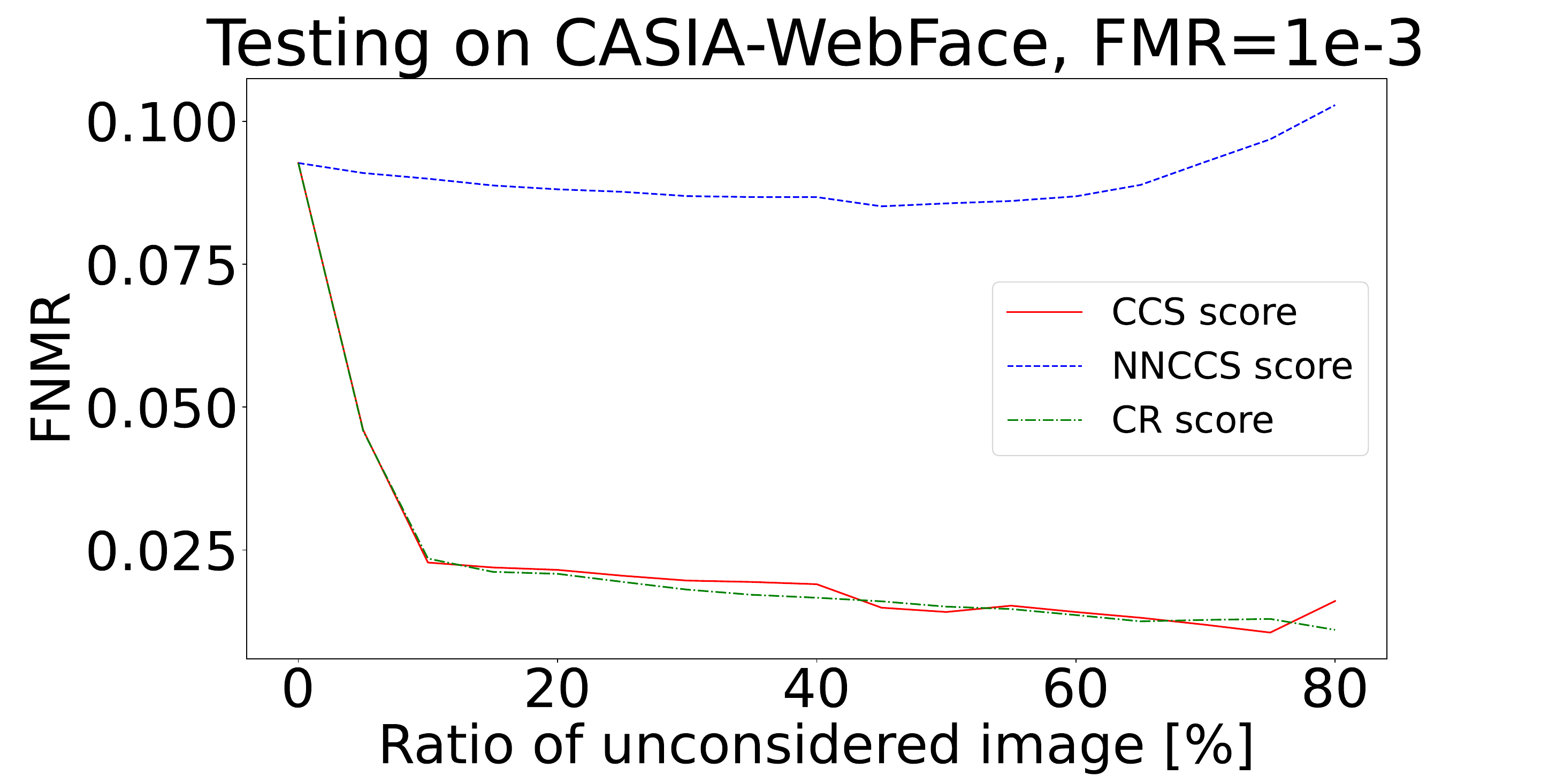}
         \caption{}
         \label{fig:erc_casio_001}
     \end{subfigure}
     \begin{subfigure}[b]{0.45\linewidth}
         \centering
   \includegraphics[width=\linewidth]{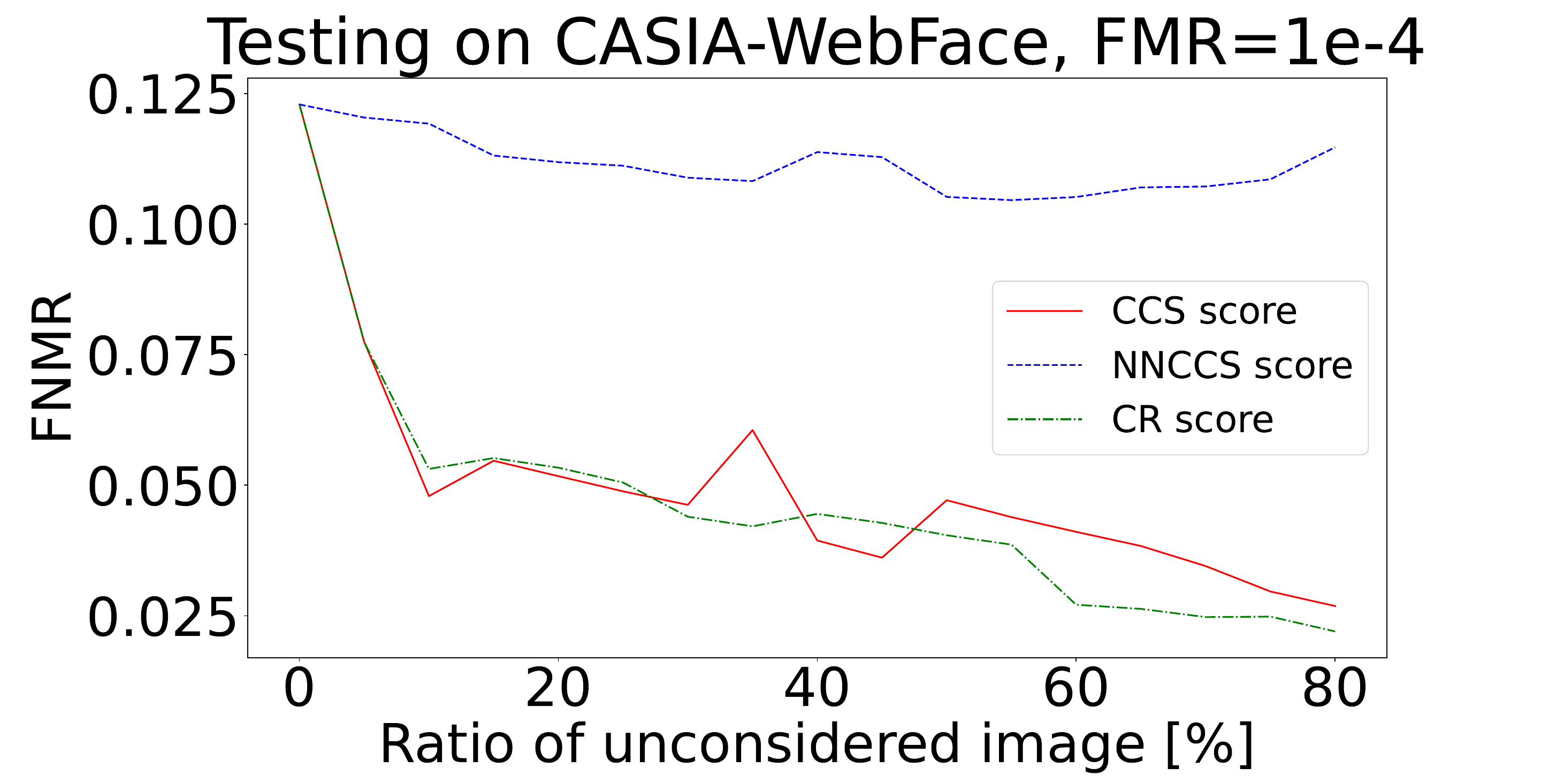}
         \caption{}
         \label{fig:erc_casio_0001}
     \end{subfigure}
   \caption{ERCs showing the verification performance as False None Match Rate (FNMR) at False Match Rate (FMR) of 1e-3 (\ref{fig:erc_casio_001}) and 1e-4 (\ref{fig:erc_casio_0001}) with CCS, NNCCS and CR as FIQ vs.~rejection ratio. This ERC plots show the effectiveness of rejecting samples with the lowest CCS and CR on the performance  }
   \label{fig:erc_casia}
\end{figure}

Figure \ref{fig:erc_casia} presents the ERC of CR, CCS, and NNCCS experimentally used as FIQ. 
An FIQ measure would cause the ERC to drop as rapidly as possible when rejecting a larger fraction of low-quality samples (moving to the right).

It can be clearly noticed in Figure \ref{fig:erc_casia} that the CCS and CR do behave as we would expect from a good performing FIQ, as the verification error value drops rapidly when rejecting low quality (low CCS and CR) samples.
It can be also observed that the CR does that more steadily when compared to CCS. 
This points out that adding the scaling term NNCCS in CR calculation can enhance the representation of the CCS as an FIQ measure, which will be clearer later when we experimentally evaluate our CR-FIQA approach in Section \ref{sec:abl}. 
As expected, the NNCCS measure by itself does not strongly act as an FIQ measure would, demonstrated by the relatively flat ERC in Figure \ref{fig:erc_casia}, as it only considers the distance to the nearest negative class.
This empirical evaluation does provide a confirming answer to the previously stated question by affirming that the CR does act as expected from an FIQ measure and thus, theoretically, one can strongly link the image properties that cause high/low CR in the FR training data to these causing high/low FIQ.



\begin{figure*}[]
    \centering
    \begin{subfigure}[b]{0.24\linewidth}
     \centering
     \includegraphics[width=\linewidth]{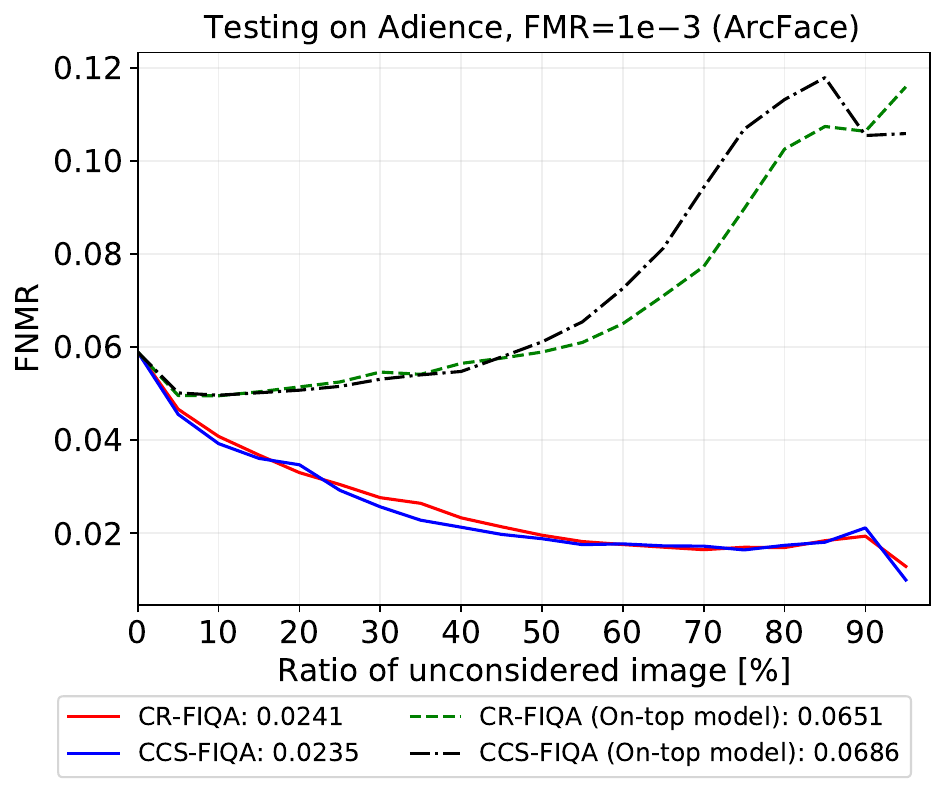}
     \caption{}
     \label{fig:as_adience}
    \end{subfigure}
        \begin{subfigure}[b]{0.24\linewidth}
     \centering
     \includegraphics[width=\linewidth]{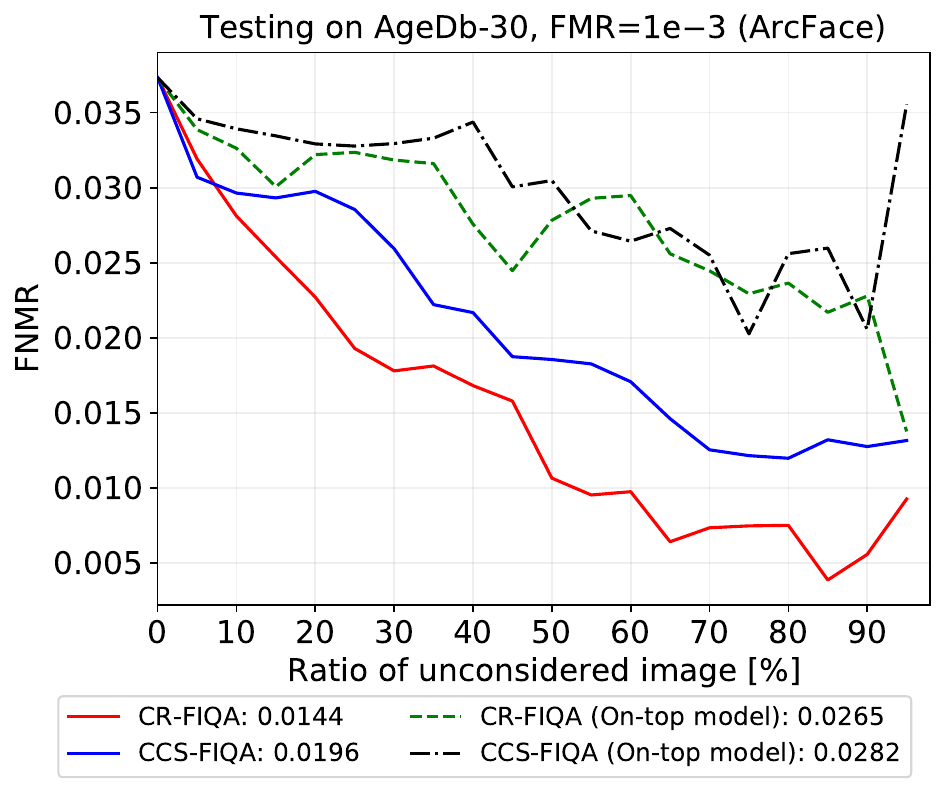}
     \caption{}
     \label{fig:as_agedb30}
    \end{subfigure}
      \begin{subfigure}[b]{0.24\linewidth}
     \centering
     \includegraphics[width=\linewidth]{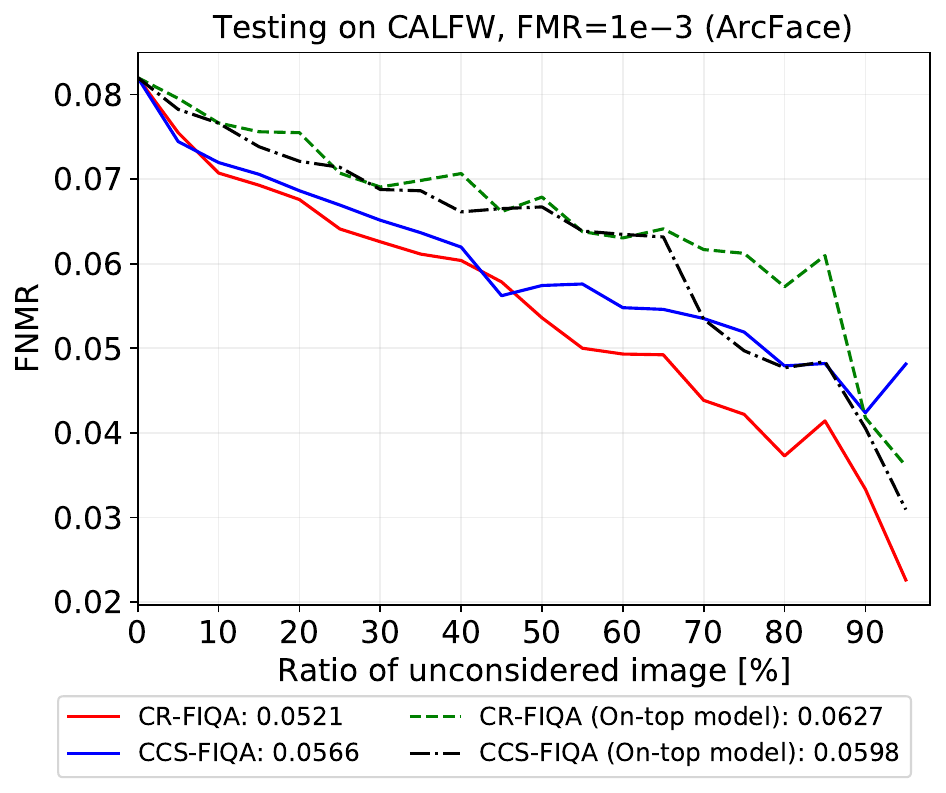}
     \caption{}
     \label{fig:as_calfw}
    \end{subfigure}
          \begin{subfigure}[b]{0.235\linewidth}
     \centering
     \includegraphics[width=\linewidth]{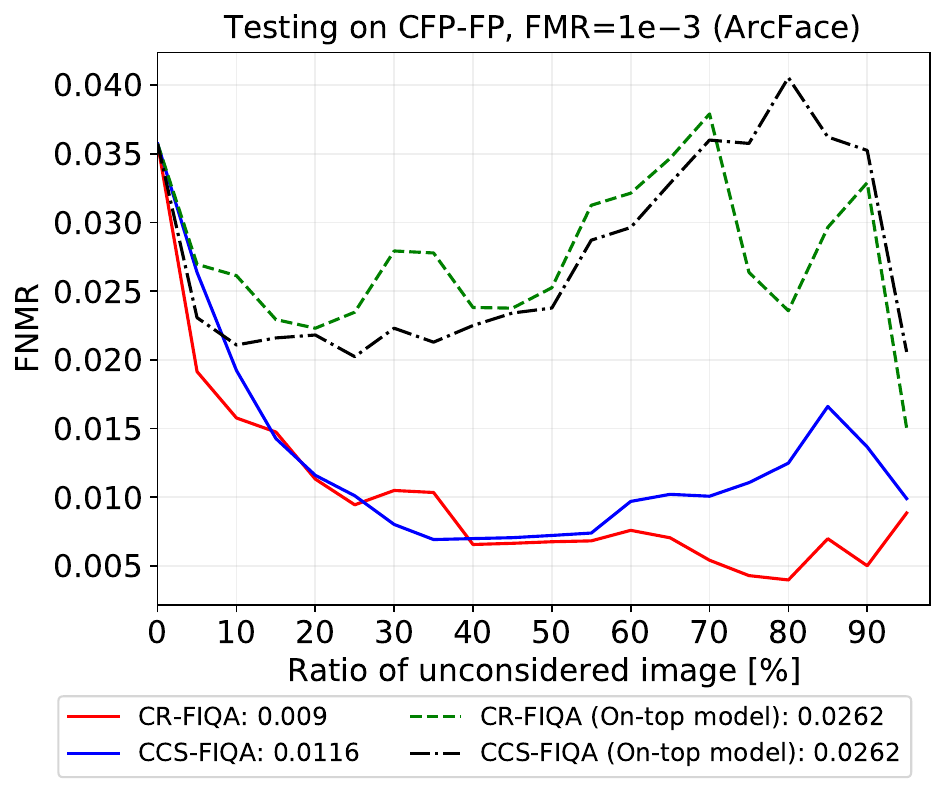}
     \caption{}
     \label{fig:as_cfpfp}
    \end{subfigure}
\caption{ERC comparison between CR-FIQA(S), CCS-FIQA(S), CR-FIQA(S) (On top) and CCS-FIQA(S) (On top). The plots show the effect of rejecting samples of lowest quality, on the verification error (FNMR at FMR1e-3). CR-FIQA(S) and CCS-FIQA(S) outperformed the on-top solutions, and CR-FIQA(S) performs generally better than CCS-FIQA(S) (curve decays faster with more rejected samples)   
}
\label{fig:ablation}
\end{figure*}

\subsection{Quality Estimation Training Paradigm}
\label{sec:Tparadigm}
In the previous section, we proved that the CR does behave as an FIQ would, and thus, it can also relate to image properties that dictate FIQ. 
However, the CR measure is only observable for samples in the FR training dataset, where the class centers are known. 
In a real case scenario, the FIQ measure should be assessed to any single image, i.e. unseen evaluation data.
Considering this, and in an effort to predict what the CR value would be for a given sample if hypothetically it was part of the FR training, we propose to simultaneously learn to predict the CR from the training dataset while optimizing the class centers (typical FR training) during the training phase, i.e. the CR-FIQA model. 
To enable this, we add a single regression layer to the FR model. 
The input of the regression layer is a feature embedding $x_{i}$ and the output is an estimation of the CR. 
The output of this regression layer is used later to predict the FIQ score of the unseen sample, e.g. from the evaluation dataset. 
Thus, we capture the properties that make the CR high/low to predict the FIQ of any given sample.
Towards this goal, during the training phase, the model (in Figure \ref{fig:workflow}) has two learning objectives: a) It is trained to optimize the distance between the samples and the class centers using ArcFace loss defined in Equation \ref{eq:arc}.
b) It is trained to predict the internal network observation, CR, using Smooth L1-Loss  \cite{smoothl1} applied between the output of the regression layer (P) and the CR calculated as in Equation \ref{eq:cr}.  
Smooth L1-loss can be interpreted as a combination of L1 and L2-losses by defining a threshold $\beta$ that changes between them \cite{smoothl1}. 
Our choice for smooth-l1 loss is based on: 1) It is less sensitive to outliers than l2.
The derivative of L2 loss increases when the difference between the prediction and ground-truth label is increased, making the derivative of loss values large at the early stage of the training, leading to unstable training. 
Additionally, L2 loss can easily generate gradient explosion \cite{smoothl1} when there are outliers in the training data. 
2) L1 loss can lead to stable training. However, the absolute values of the difference between prediction and ground truth are small, especially in the later stage of the training. Therefore, the model accuracy can hardly be improved at a later stage of the training as the loss function will fluctuate around a stable value.
Combining L1 and L2 as in Smooth L1-loss avoids gradient explosion, which might be caused by L2 and facilitates better convergence than L1.
The loss leading to the second objective is then given as: 
\begin{equation}
\label{eq:cr}
\resizebox{\linewidth}{!}{$
   \mathcal{L}_{CR}=  \frac{1}{N}\sum\limits_{i \in N}
    \begin{cases}
    \frac{ 0.5 \times (CR_{x_i}-P_{i})^2}{\beta} & \text{if } |CR_{x_i}-P_{i}|<\beta \\
     |CR_{x_i}-P_{i}| -0.5 \times \beta & \text{otherwise }
    \end{cases}$}.
\end{equation}
The final loss combining both objectives for training our CR-FIQA model is defined as follows: 
\begin{equation}
\label{eq:cr_fiqa}
   \mathcal{L}= \mathcal{L}_{Arc} + \lambda \times \mathcal{L}_{CR},
\end{equation}
where $\lambda$ is a hyper-parameter used to control the balance between the two losses. At the beginning of model training, the value range of $\mathcal{L}_{CR}$ is very small ($\leq 2$) in comparison to $\mathcal{L}_{Arc}$ ($\sim 45$).
Setting $\lambda$ to a small value, the model will only focus on $\mathcal{L}_{Arc}$. 
Besides, setting $\lambda$ to a large value, i.e.~$>10$, we observed that the model did not converge.
Therefore, we set $\lambda$ to $10$ in all the experiments in this paper.



\section{Experimental Setup}
\label{sec:exp}

\textbf{Implementation Details}
We demonstrate our proposed CR-FIQA under two protocols (small and large) based on the training dataset and the training model architecture.
We utilize widely used architectures in the SOTA FR solutions, ResNet100 and ResNet50  \cite{DBLP:conf/cvpr/HeZRS16}, both modified as described in Section \ref{sec:Tparadigm}. 
For the small protocol, we utilize ResNet50 and the CASIA-WebFace \cite{casia_webface} training data (noted as CR-FIQA(S)) and for the large protocol, we utilize ResNet100 and the MS1MV2 \cite{guo2016ms,deng2019arcface} training data (noted as CR-FIQA(L)). 
The MS1MV2 is a refined version of the MS-Celeb-1M \cite{guo2016ms} by \cite{deng2019arcface} containing 5.8M images of 85K identities. 
The CASIA-WebFace contains 0.5m images of 10K identities \cite{casia_webface}. 
We follow the ArcFace training setting \cite{deng2019arcface} to set the scale parameter $s$ to 64 and the margin $m$ to 0.5.  
We set the mini-batch size to 512.
All models are trained with Stochastic Gradient Descent (SGD) optimizer with an initial learning rate of 1e-1. 
During the training, we use random horizontal flipping with a probability of 0.5 for data augmentation.
We set the momentum to 0.9 and the weight decay to 5e-4. 
For CR-FIQA(S), the learning rate is divided by 10 at 20K and at 28K training iterations, following \cite{deng2019arcface}. 
The training is stopped after 32K iterations. 
For CR-FIQA(L), the learning rate is divided by 10 at 100K and 160K training iterations, following \cite{deng2019arcface}. The training is stopped after 180K iterations.
All the images in evaluation and training datasets are aligned and cropped to $112 \times 112$, as described in \cite{deng2019arcface}.
All the training and testing images are normalized to have pixel values between -1 and 1.
Both models are trained using the loss defined in Equation \ref{eq:cr_fiqa}. 

\textbf{Evaluation Benchmarks} 
We reported the achieved results on eight different benchmarks: Labeled Faces in the Wild (LFW) \cite{LFWTech}, AgeDB-30 \cite{agedb}, Celebrities in Frontal-Profile in the Wild (CFP-FP) \cite{cfp-fp}, Cross-age LFW (CALFW) \cite{CALFW}, Adience \cite{Adience}, Cross-Pose LFW (CPLFW)  \cite{CPLFWTech}, Cross-Quality LFW (XQLFW) \cite{XQLFW}, IARPA Janus Benchmark–C (IJB-C) \cite{ijbc}. 
These benchmarks are chosen to provide a wide comparison to SOTA FIQA algorithms and give an insight into the CR-FIQA generalizability. 

\textbf{Evaluation Metric}
We evaluate the FIQA by plotting ERCs \cite{GT07}. The ERC is a widely used representation of the FIQA performance \cite{GT07,NISTQuaity} by demonstrating the effect of rejecting a fraction face images, of the lowest quality, on face verification performance in terms of False None Match Rate \cite{iso_metric} (FNMR) at a specific threshold calculated at fixed False Match Rate \cite{iso_metric} (FMR).  The ERC curves for all benchmarks are plotted at two fixed FMRs, 1e-3 (as recommended for border control operations by Frontex \cite{frontex2015best}) and 1e-4 (the latter is provided in the supplementary material). 
We also report the Area under the Curve (AUC) of the ERC, to provide a quantitative aggregate measure of verification performance across all rejection ratios.

Additionally, motivated by evaluating the FIQ as a weighting term for face embedding \cite{SDDFIQA,PFE_FIQA}, we follow the IJB-C 1:1 mixed verification benchmark \cite{ijbc} by weighting the frames such that all frames belonging to the same subject within a video have a combined weight equal to a single still image as described in IJB-C benchmark \cite{ijbc}.
We do that by using the CR-FIQA quality scores as well as all SOTA methods.
We report the verification performance of IJB-C as true acceptance rates (TAR) at false acceptance rates (FAR) of 1e-4, 1e-5, and 1e-6, as defined in \cite{ijbc}. 

\textbf{Face Recognition Models}
We utilize four different SOTA FR models to report the verification performance at different quality rejection rate to inspect the generalizability of FIQA over FR solutions. The FR models are ArcFace \cite{deng2019arcface}, ElasticFace (ElasticFace-Arc) \cite{elasticface}, MagFace \cite{MagFace}, and CurricularFace \cite{curricularFace}. All models process $112 \times 112$ aligned and cropped image to produce 512-D feature embedding.
We used the officially released pretrained  ResNet-100 models trained on MS1MV2 released by the four FR solutions.
Although, the presented solution in this paper does not aim, and is not presented as, a solution to extract face embeddings, but rather an FIQA solution,  we opted to evaluate CR-FIQA(L) backbone as a FR model on mainstream FR benchmarks for sake of providing complete experiment evaluation and probe the possibility of simultaneously using it as both FIQA and FR model. The evaluation results of CR-FIQA(L) backbone as a FR model are provided in the supplementary material.


\textbf{Baseline}
We compare our CR-FIQA approach with nine quality assessment methods. Three are general IQA methods that have been proven in \cite{BiyingWACV} to correlate well to face utility i.e. BRISQUE \cite{BRISQE_IQA}, RankIQA \cite{liu2017rankiqa}, and DeepIQA \cite{DEEPIQ_IQA}, and six are SOTA face-specific FIQA methods, namely RankIQ \cite{RANKIQ_FIQA}, PFE \cite{PFE_FIQA}, SER-FIQ \cite{SERFIQ}, FaceQnet (v1 \cite{faceqnetv1}) \cite{hernandez2019faceqnet,faceqnetv1}, MagFace \cite{MagFace}, and SDD-FIQA \cite{SDDFIQA}, all as officially released in the respective works.

\begin{table*}[ht!]
\begin{center}
\caption{The AUCs of ERC achieved by our CR-FIQA and the SOTA methods under different experimental settings. 
CR-FIQA achieved the best performance (lowest AUC) in almost all settings. 
On XQLFW, the SER-FIQ (marked with *) is used for the sample selection of the XQLFW benchmark.
The best result for each experimental setting is in bold and the second-ranked one is in italic. The notions of $1e-3$ and $1e-4$ indicate the value of the fixed FMR at which the ERC curve (FNMR vs.~reject) was calculated at }
\label{tab:erc}
\resizebox{0.93\textwidth}{!}{
\begin{tabular}{|c|cl|ll|ll|ll|ll|ll|ll|ll|ll|}
\hline
 \multirow{2}{*}{FR} & \multicolumn{2}{c|}{\multirow{2}{*}{Method}} & \multicolumn{2}{l|}{Adience\cite{Adience}} & \multicolumn{2}{l|}{AgeDB-30\cite{agedb}} & \multicolumn{2}{l|}{CFP-FP\cite{cfp-fp}} & \multicolumn{2}{l|}{LFW\cite{LFWTech}} & \multicolumn{2}{l|}{CALFW\cite{CALFW}} & \multicolumn{2}{l|}{CPLFW\cite{CPLFWTech}} & \multicolumn{2}{l|}{XQLFW\cite{XQLFW}} & \multicolumn{2}{l|}{IJB-C\cite{ijbc}}\\  \cline{4-19}
 & &  & $1e{-3}$ & $1e{-4}$ & $1e{-3}$ & $1e{-4}$  & $1e{-3}$ & $1e{-4}$  & $1e{-3}$ & $1e{-4}$  & $1e{-3}$ & $1e{-4}$ & $1e^{-3}$ & $1e{-4}$ & $1e{-3}$ & $1e{-4}$ & $1e{-3}$ & $1e{-4}$ \\ \hline
 \hline 
 \multirow{12}{*}{\rotatebox[origin=c]{90}{ArcFace\cite{deng2019arcface}}} & \multirow{3}{*}{\rotatebox[origin=c]{90}{IQA}} & \multicolumn{1}{|l|}{BRISQUE\cite{BRISQE_IQA}} &0.0565 & 0.1285 &0.0400 & 0.0585 &0.0343 & 0.0433 &0.0043 & 0.0049 &0.0755 & 0.0813 &0.2558 & 0.3037 &0.6680 & 0.7122 & 0.0381 & 0.0656 \\ 
 & & \multicolumn{1}{|l|}{RankIQA\cite{liu2017rankiqa}} &0.0400 & 0.0933 &0.0372 & 0.0523 &0.0301 & 0.0384 &0.0039 & 0.0045 &0.0846 & 0.0915 &0.2437 & 0.2969 &0.6584 & 0.7039 & 0.0385 & 0.0640  \\ 
 & & \multicolumn{1}{|l|}{DeepIQA\cite{DEEPIQ_IQA}} &0.0568 & 0.1372 &0.0403 & 0.0523 &0.0238 & 0.0292 &0.0049 & 0.0056 &0.0793 & 0.0850 &0.2309 & 0.2856 &0.5958 & 0.6458 & 0.0383 & 0.0640 \\ \cline{2-19} 
 & \multirow{8}{*}{\rotatebox[origin=c]{90}{FIQA}} & \multicolumn{1}{|l|}{RankIQ\cite{RANKIQ_FIQA} }&0.0353 & 0.0873 &0.0322 & 0.0420 &0.0152 & 0.0260 &\textit{0.0018} & \textit{0.0024} &0.0608 & 0.0672 &0.0633 & 0.0848 &0.2789 & 0.3332 & 0.0227 & 0.0342 \\ 
 & & \multicolumn{1}{|l|}{PFE\cite{PFE_FIQA}} &0.0212 & 0.0428 &0.0172 & 0.0226 &0.0092 & 0.0129 &0.0023 & 0.0028 &0.0647 & 0.0681 & 0.0450 & 0.0638 &0.2302 & 0.2710 & 0.0176 & 0.0248  \\ 
 & & \multicolumn{1}{|l|}{SER-FIQ\cite{SERFIQ}} &0.0223 & 0.0434 &0.0167 & 0.0223 & \textit{0.0065} & \textit{0.0103} & 0.0023 & 0.0028 & 0.0595 & 0.0627 &\textit{0.0389} & 0.0584 & \textbf{0.1812}$^{*}$ & \textbf{0.2295}$^{*}$ & \textit{0.0161} & \textit{0.0241} \\ 
 & & \multicolumn{1}{|l|}{FaceQnet\cite{hernandez2019faceqnet,faceqnetv1}} & 0.0346 & 0.0734 &0.0197 & 0.0245 &0.0240 & 0.0273 &0.0022 & 0.0027 &0.0774 & 0.0822 &0.1504 & 0.1751 &0.5829 & 0.6136 & 0.0270 & 0.0376\\ 
 & & \multicolumn{1}{|l|}{MagFace\cite{MagFace}} & \textit{0.0207} & \textit{0.0425} & \textit{0.0156} & 0.0198 &0.0073 & 0.0105 & \textbf{0.0016} & \textbf{0.0021} & \textit{0.0568} & \textit{0.0602} &0.0492 & 0.0642 &0.4022 & 0.4636 & 0.0171 & 0.0254 \\ 
 & & \multicolumn{1}{|l|}{SDD-FIQA\cite{SDDFIQA}} &0.0248 & 0.0562 &0.0186 & 0.0206 &0.0122 & 0.0193 &0.0021 & 0.0027 &0.0641 & 0.0698 &0.0517 & 0.0670 &0.3090 & 0.3561 & 0.0186 & 0.0270\\ \cline{3-19}
 & & \multicolumn{1}{|l|}{CR-FIQA(S)(Our)} & 0.0241 & 0.0517 & \textbf{0.0144} & \textbf{0.0187} &0.0090 & 0.0145 &0.0020 & 0.0025 & \textbf{0.0521} & \textbf{0.0554} &0.0391 & \textit{0.0567} &0.2377 & 0.2740 & 0.0171 & 0.0250 \\ 
 & & \multicolumn{1}{|l|}{CR-FIQA(L)(Our)} & \textbf{0.0204} & \textbf{0.0353} &0.0159 & \textit{0.0189} & \textbf{0.0050} & \textbf{0.0082} &0.0023 & 0.0029 &0.0616 & 0.0632 & \textbf{0.0360} & \textbf{0.0515} & \textit{0.2084} & \textit{0.2441} & \textbf{0.0138} & \textbf{0.0207} \\
 \hline \hline 
 \multirow{12}{*}{\rotatebox[origin=c]{90}{ElasticFace\cite{elasticface}}} & \multirow{3}{*}{\rotatebox[origin=c]{90}{IQA}} & \multicolumn{1}{|l|}{BRISQUE\cite{BRISQE_IQA}} &0.0644 & 0.1184 &0.0375 & 0.0403 &0.0281 & 0.0372 &0.0034 & 0.0047 &0.0726 & 0.0747 &0.2641 & 0.4688 &0.6343 & 0.6964 & 0.0357 & 0.0621\\ 
 & & \multicolumn{1}{|l|}{RankIQA\cite{liu2017rankiqa}} &0.0433 & 0.0862 &0.0374 & 0.0436 &0.0269 & 0.0318 &0.0033 & 0.0045 &0.0810 & 0.0835 &0.2325 & 0.4306 &0.6189 & 0.6856 & 0.0366 & 0.0599 \\ 
 & & \multicolumn{1}{|l|}{DeepIQA\cite{DEEPIQ_IQA}} &0.0645 & 0.1203 &0.0384 & 0.0411 &0.0191 & 0.0256 &0.0043 & 0.0056 &0.0756 & 0.0772 &0.2401 & 0.4541 &0.5400 & 0.5832 & 0.0379 & 0.0590 \\ \cline{2-19} 
 &\multirow{8}{*}{\rotatebox[origin=c]{90}{FIQA}} & \multicolumn{1}{|l|}{RankIQ\cite{RANKIQ_FIQA} }& 0.0400 & 0.0777 &0.0309 & 0.0337 &0.0149 & 0.0180 & \textbf{0.0013} & \textbf{0.0020} &0.0598 & 0.0614 &0.0581 & 0.0727 &0.2468 & 0.2776 & 0.0226 & 0.0334 \\ 
 & & \multicolumn{1}{|l|}{PFE\cite{PFE_FIQA}} & \textit{0.0222} & \textit{0.0381} &0.0163 & 0.0172 &0.0088 & 0.0113 &0.0018 & 0.0025 &0.0628 & 0.0643 &0.0419 & 0.0895 &0.2112 & 0.2436 & 0.0171 & 0.0247 \\ 
 & & \multicolumn{1}{|l|}{SER-FIQ\cite{SERFIQ} } &0.0240 & 0.0417 &0.0163 & 0.0179 &\textit{0.0061} & \textit{0.0085} &0.0021 & 0.0028 &0.0574 & 0.0590 &0.0387 & 0.0513 & \textbf{0.1576}$^{*}$ & \textbf{0.1868}$^{*}$ & \textit{0.0156} & \textit{0.0235} \\ 
 & & \multicolumn{1}{|l|}{FaceQnet\cite{hernandez2019faceqnet,faceqnetv1}} &0.0369 & 0.0667 &0.0194 & 0.0207 &0.0227 & 0.0247 &0.0021 & 0.0026 &0.0763 & 0.0777 &0.1420 & 0.2880 &0.5549 & 0.5844 & 0.0263 & 0.0370 \\ 
 & & \multicolumn{1}{|l|}{MagFace\cite{MagFace}} & 0.0225 & 0.0385 &0.0150 & \textbf{0.0158} &0.0069 & 0.0095 & \textit{0.0014} & \textit{0.0021} & \textit{0.0553} & \textit{0.0563} & 0.0474 & 0.0597 &0.3973 & 0.4282 & 0.0166 & 0.0243 \\ 
 & & \multicolumn{1}{|l|}{SDD-FIQA\cite{SDDFIQA}} &0.0277 & 0.0512 &0.0187 & 0.0200 &0.0098 & 0.0118 &0.0019 & 0.0027 &0.0624 & 0.0638 &0.0493 & 0.0634 &0.3052 & 0.3562 & 0.0183 & 0.0266 \\ \cline{3-19}
 & & \multicolumn{1}{|l|}{CR-FIQA(S)(Our)} &0.0257 & 0.0465 & \textbf{0.0146} & 0.0160 &0.0070 & 0.0096 &0.0015 & 0.0022 & \textbf{0.0509} & \textbf{0.0522} & \textit{0.0383} & \textit{0.0502} &0.2093 & 0.2835 & 0.0167 & 0.0244 \\ 
 & & \multicolumn{1}{|l|}{CR-FIQA(L)(Our)} & \textbf{0.0214} & \textbf{0.0357} & \textit{0.0149} & \textit{0.0159} & \textbf{0.0045} & \textbf{0.0065} &0.0018 & 0.0025 &0.0594 & 0.0608 & \textbf{0.0350} & \textbf{0.0462} &\textit{0.1798} & \textit{0.2060} & \textbf{0.0135} & \textbf{0.0203} \\ 
 \hline \hline 
 \multirow{12}{*}{\rotatebox[origin=c]{90}{MagFace\cite{MagFace}}} & \multirow{3}{*}{\rotatebox[origin=c]{90}{IQA}} & \multicolumn{1}{|l|}{BRISQUE\cite{BRISQE_IQA}} &0.0594 & 0.1308 &0.0442 & 0.0799 &0.0422 & 0.0589 &0.0043 & 0.0058 &0.0758 & 0.0788 &0.4649 & 0.6809 &0.6911 & 0.7229 & 0.0462 & 0.0787 \\ 
 & & \multicolumn{1}{|l|}{RankIQA\cite{liu2017rankiqa}} &0.0407 & 0.0889 &0.0370 & 0.0681 &0.0369 & 0.0543 &0.0041 & 0.0056 &0.0829 & 0.0857 &0.3251 & 0.6475 &0.6706 & 0.7046 & 0.0461 & 0.0750 \\ 
 & & \multicolumn{1}{|l|}{DeepIQA\cite{DEEPIQ_IQA}} &0.0571 & 0.1302 &0.0417 & 0.0721 &0.0322 & 0.0545 &0.0048 & 0.0059 &0.0787 & 0.0809 &0.3672 & 0.6632 &0.6162 & 0.6519 & 0.0474 & 0.0765 \\ \cline{2-19} 
 & \multirow{8}{*}{\rotatebox[origin=c]{90}{FIQA}} & \multicolumn{1}{|l|}{RankIQ\cite{RANKIQ_FIQA}}&0.0359 & 0.0837 &0.0361 & 0.0531 &0.0213 & 0.0332 & \textit{0.0019} & \textit{0.0027} &0.0602 & 0.0629 &0.0659 & 0.1642 &0.3076 & 0.3475 & 0.0269 & 0.0383 \\ 
 & & \multicolumn{1}{|l|}{PFE\cite{PFE_FIQA}} &0.0215 & 0.0423 &0.0192 & 0.0317 &0.0107 & 0.0138 &0.0023 & 0.0029 &0.0640 & 0.0652 &0.0449 & 0.1435 &0.2615 & 0.2926 & 0.0200 & 0.0283  \\ 
 &  & \multicolumn{1}{|l|}{SER-FIQ\cite{SERFIQ} } &0.0233 & 0.0451 &0.0185 & 0.0293 &\textit{0.0080} & 0.0139 &0.0025 & 0.0033 &0.0590 & 0.0607 & \textit{0.0397} & \textit{0.0821} & \textbf{0.2139}$^*$ & \textbf{0.2562}$^*$ & \textit{0.0189} & \textit{0.0270}\\ 
 & & \multicolumn{1}{|l|}{FaceQnet\cite{hernandez2019faceqnet,faceqnetv1}} &0.0365 & 0.0720 &0.0217 & 0.0314 &0.0271 & 0.0351 &0.0022 & \textit{0.0027} &0.0763 & 0.0773 &0.2988 & 0.5218 &0.6016 & 0.6210 & 0.0305 & 0.0422 \\ 
 & & \multicolumn{1}{|l|}{MagFace\cite{MagFace}} & \textit{0.0212} & \textit{0.0417} & \textbf{0.0159} & 0.0247 &0.0085 & 0.0129 &\textbf{0.0017} & \textbf{0.0022} & \textit{0.0562} & \textit{0.0578} &0.0506 & 0.0887 &0.4478 & 0.4900 & 0.0195 & 0.0279 \\ 
 & & \multicolumn{1}{|l|}{SDD-FIQA\cite{SDDFIQA}} &0.0253 & 0.0562 &0.0216 & 0.0305 &0.0146 & 0.0201 &0.0021 & \textit{0.0027} &0.0643 & 0.0657 &0.0525 & 0.1188 &0.3404 & 0.3928 & 0.0215 & 0.0307 \\ \cline{3-19}
 & & \multicolumn{1}{|l|}{CR-FIQA(S)(Our)} &0.0244 & 0.0507 &\textit{0.0165} & \textbf{0.0234} & 0.0102 & \textit{0.0121} &0.0020 & 0.0028 & \textbf{0.0516} & \textbf{0.0528} & 0.0409 & 0.0840 &0.2670 & 0.3336 & 0.0198 & 0.0284 \\ 
 & & \multicolumn{1}{|l|}{CR-FIQA(L)(Our)} & \textbf{0.0211} & \textbf{0.0372} &0.0174 & \textit{0.0235} &\textbf{0.0062} & \textbf{0.0080} &0.0023 & 0.0028 &0.0614 & 0.0628 & \textbf{0.0374} & \textbf{0.0679} & \textit{0.2369} & \textit{0.2839} & \textbf{0.0162} & \textbf{0.0236} \\
 \hline \hline 
 \multirow{12}{*}{\rotatebox[origin=c]{90}{CurricularFace\cite{curricularFace}}} & \multirow{3}{*}{\rotatebox[origin=c]{90}{IQA}} & \multicolumn{1}{|l|}{BRISQUE\cite{BRISQE_IQA}} &0.0502 & 0.1095 &0.0433 & 0.0491 &0.0323 & 0.0357 &0.0041 & 0.0047 &0.0755 & 0.0784 &0.2709 & 0.5057 &0.6146 & 0.6336 & 0.0362 & 0.0589 \\ 
 & & \multicolumn{1}{|l|}{RankIQA\cite{liu2017rankiqa}} &0.0359 & 0.0752 &0.0394 & 0.0510 &0.0298 & 0.0356 &0.0039 & 0.0045 &0.0806 & 0.0865 &0.2346 & 0.4654 &0.5900 & 0.6212 & 0.0361 & 0.0556 \\ 
 & & \multicolumn{1}{|l|}{DeepIQA\cite{DEEPIQ_IQA}} &0.0492 & 0.1070 &0.0407 & 0.0476 &0.0227 & 0.0278 &0.0050 & 0.0056 &0.0764 & 0.0786 &0.2488 & 0.4961 &0.5165 & 0.5526 & 0.0376 & 0.0571 \\ \cline{2-19} 
 &\multirow{8}{*}{\rotatebox[origin=c]{90}{FIQA}} & \multicolumn{1}{|l|}{RankIQ\cite{RANKIQ_FIQA} }&0.0314 & 0.0715 &0.0365 & 0.0417 &0.0186 & 0.0249 & \textit{0.0018} & \textit{0.0024} &0.0590 & 0.0640 &0.0541 & 0.0730 &0.2449 & 0.2880 & 0.0220 & 0.0320 \\ 
 & & \multicolumn{1}{|l|}{PFE\cite{PFE_FIQA}} & \textbf{0.0198} & 0.0365 &0.0197 & 0.0227 &0.0100 & 0.0134 &0.0024 & 0.0028 &0.0630 & 0.0657 &0.0402 & 0.0983 &0.1982 & 0.2220 & 0.0170 & 0.0238 \\ 
 & & \multicolumn{1}{|l|}{SER-FIQ\cite{SERFIQ}} &0.0211 & 0.0381 &0.0167 & \textbf{0.0193} & \textit{0.0074} & \textit{0.0111} &0.0025 & 0.0030 &0.0587 & 0.0610 &0.0356 & 0.0520 & \textbf{0.1558}$^*$ & \textbf{0.1866}$^*$ & \textit{0.0153} & \textit{0.0228}\\ 
 & & \multicolumn{1}{|l|}{FaceQNet\cite{hernandez2019faceqnet,faceqnetv1}} &0.0326 & 0.0626 &0.0221 & 0.0267 &0.0226 & 0.0274 &0.0022 & 0.0027 &0.0767 & 0.0799 &0.1384 & 0.3229 &0.5035 & 0.5411 & 0.0259 & 0.0354 \\ 
 & & \multicolumn{1}{|l|}{MagFace\cite{MagFace}} &\textit{0.0200} & \textit{0.0364} &0.0167 & 0.0195 &0.0078 & 0.0111 & \textbf{0.0016} & \textbf{0.0021} & \textit{0.0563} & \textit{0.0590} &0.0449 & 0.0607 &0.3758 & 0.4178 & 0.0163 & 0.0232 \\ 
 & & \multicolumn{1}{|l|}{SDD-FIQA\cite{SDDFIQA}} &0.0230 & 0.0462 &0.0219 & 0.0254 &0.0138 & 0.0185 &0.0021 & 0.0027 &0.0637 & 0.0675 &0.0465 & 0.0671 &0.2649 & 0.3053 & 0.0178 & 0.0254\\ \cline{3-19}
 & & \multicolumn{1}{|l|}{CR-FIQA(S)(Our)} &0.0227 & 0.0446 & \textbf{0.0156} & \textit{0.0198} &0.0097 & 0.0148 &0.0020 & 0.0025 &\textbf{0.0513} & \textbf{0.0534} & \textit{0.0340} & \textit{0.0501} &0.2101 & 0.2470 & 0.0165 & 0.0234 \\ 
 & & \multicolumn{1}{|l|}{CR-FIQA(L)(Our)} &\textbf{0.0198} & \textbf{0.0336} &\textit{0.0162} & 0.0200 & \textbf{0.0054} & \textbf{0.0080} &0.0023 & 0.0029 &0.0605 & 0.0618 &\textbf{0.0324} & \textbf{0.0462} & \textit{0.1716} & \textit{0.2318} & \textbf{0.0134} & \textbf{0.0194} \\ 
 \hline 
\end{tabular}}
\end{center}
\end{table*}

\begin{table*}[h!]
\centering
\caption{Verification performance on the IJB-C (1:1 mixed verification) \cite{ijbc}. CR-FIQA outperformed SOTA methods under all settings }
\label{tab:ijbc_ver}
\resizebox{0.93\linewidth}{!}{%
\begin{tabular}{|cc|ccc|ccc|ccc|ccc|}
\hline
\multicolumn{2}{|c|}{\multirow{3}{*}{Quality Estimation}} &
\multicolumn{12}{c|}{1:1 mixed Verification: TAR (\%) at} \\ \cline{3-14}
& & \multicolumn{3}{c|}{ArcFace\cite{deng2019arcface}} & \multicolumn{3}{c|}{ElasticFace \cite{elasticface}} & \multicolumn{3}{c|}{MagFace \cite{MagFace}} & \multicolumn{3}{c|}{CurricularFace\cite{curricularFace}} \\ \cline{3-14}
& & {FAR=1e-6} & {FAR=1e-5} & {FAR=1e-4} & {FAR=1e-6} & {FAR=1e-5} & {FAR=1e-4} & {FAR=1e-6} & {FAR=1e-5} & {FAR=1e-4} & {FAR=1e-6} & {FAR=1e-5} & {FAR=1e-4}\\ \hline \hline
& \multicolumn{1}{|l|}{-} & \textit{89.85} & 94.47  & 96.28  & 89.15 & 94.54 & 96.49 & 85.67  & 93.08 & 96.65  & 90.46 & 94.89 & 96.58   \\ \hline
\multirow{3}{*}{\rotatebox[origin=c]{90}{IQA}} & \multicolumn{1}{|l|}{BRISQUE\cite{BRISQE_IQA}} & 86.65  & 93.62  & 95.98 & 85.68   & 93.51  & 95.65 & 81.11 & 90.64 & 94.82 & 88.16 & 93.98 & 96.29  \\
& \multicolumn{1}{|l|}{RankIQA\cite{liu2017rankiqa}} & 86.37 & 93.61 & 95.83  & 86.71 & 93.46 & 96.00 & 80.78 & 90.75 & 94.86  & 88.16  & 94.11 & 96.22 \\
& \multicolumn{1}{|l|}{DeepIQA\cite{DEEPIQ_IQA}} & 81.97 & 91.64  & 94.67 & 78.93  & 91.59  & 94.81  & 73.53 & 86.34  & 92.90  & 82.65 & 92.04  & 95.00 \\ \hline
\multirow{8}{*}{\rotatebox[origin=c]{90}{FIQA}} & \multicolumn{1}{|l|}{RankIQ\cite{RANKIQ_FIQA}} & 88.78 & 94.42   & 96.20   & 88.88  & 94.64  & 96.45  & 85.63  & 92.66  & 95.70  & 90.00 & 94.93 & 96.53 \\
& \multicolumn{1}{|l|}{PFE\cite{PFE_FIQA}} & 89.50 & 94.51& 96.31 & 89.10 & 94.67 & 96.51 & 84.93 & 92.44 & 95.60  & 90.36 & 95.04 & 96.54 \\
& \multicolumn{1}{|l|}{SER-FIQ\cite{SERFIQ}} & 89.74  & 94.65  & 96.32& \textit{90.05}  & 94.79 & 96.57 & 86.02 & 93.35  & 95.80 & 90.66 & 95.11  & 96.58 \\
& \multicolumn{1}{|l|}{FaceQNet\cite{hernandez2019faceqnet,faceqnetv1}} & 87.87 & 94.04 & 96.12 & 86.26  & 94.09  & 96.25  & 82.91 & 90.56 & 95.03 & 89.61  & 94.65 & 96.36 \\
& \multicolumn{1}{|l|}{MagFace\cite{MagFace}} & 89.49  & 94.41  & 96.22  & 89.37 & 94.69 & 96.46  & 85.75 & 92.71 & 95.54  & 90.34 & 95.02  & 96.50 \\
& \multicolumn{1}{|l|}{SDD-FIQA\cite{SDDFIQA}} & 89.39 & 94.61 & 96.34 & 88.07 & 94.82 & 96.49 & 84.69 & 92.83 & 95.73 & 89.91 & \textit{95.12} & \textit{96.63} \\ \cline{2-14}
& \multicolumn{1}{|l|}{CR-FIQA(S)(Our)} & 89.59 & \textbf{94.78} & \textit{96.35} & \textbf{90.30 } & \textbf{94.97} & \textbf{96.63} & \textit{86.45}  & \textit{93.48} & \textbf{95.95} & \textbf{90.82} & \textbf{95.13} & \textbf{96.64} \\
& \multicolumn{1}{|l|}{CR-FIQA(L)(Our)} & \textbf{90.16} & \textit{94.75} & \textbf{96.36} & 90.00 & \textit{94.92}& \textit{96.58}  & \textbf{87.12}  &\textbf{93.67} & \textit{95.90} & \textit{90.79} & \textit{95.12} & 96.58 \\ \hline
\end{tabular}
}
\end{table*}

\section{Ablation Studies}
\label{sec:abl}
This section provides experimental proof of the two main design choices in CR-FIQA.

\textbf{Does CR-FIQA benefit from the NNCCS scaling term?}
To answer this, we conducted additional experiments using ResNet-50 model trained on CASIA-WebFace \cite{casia_webface} using the experimental setup described in Section \ref{sec:exp}. This model is noted as CCS-FIAQ(S). 
The only difference from CR-FIAQ(S) is that the CCS-FIAQ(S) is trained to learn CCS (instead of CR) by replacing the $CR_{x_i}$ in Equation \ref{eq:cr} with $CCS_{x_i}$, thus neglecting the NNCCS scaling term in the equation. 
Figure \ref{fig:ablation} presents the ERCs along with AUC  using CR-FIAQ(S) and CCS-FIAQ(S) on Adience, AgeDb-30, CALLFW and CRF-FP. The verification error, FNMR at FMR1e-3, is calculated using ArcFace FR model (described in Section \ref{sec:exp}). The ERCs and AUC values show that the reduction in the error is more evident for CR-FIAQ(S) than  CCS-FIQA(S). Thus, adding the scaling term NNCCS in CR enhanced the performance of the FIQA.



\textbf{Dose the simultaneous learning in CR-FIQA lead to better performance in comparison to on-the-top learning?}
We consider the possibility of learning to estimate CR after finalizing the FR training, in comparison to the simultaneously learning in CR-FIQA.
We conducted two additional experiments using pretrained ResNet-50 trained with ArcFace loss \cite{deng2019arcface} on CASIA-WebFace \cite{casia_webface}. 
Specifically, we add an additional single regression layer to this pretrained model. 
We freeze the weights of the pretrained model and train only the regression layer to learn the internal network observation of the pretrained model using only the $\mathcal{L_{CR}}$ (Equation \ref{eq:cr}). 
Using this setting, we present two instances, CR-FIQA(S) (On top) and CCS-FIQA(S) (On top), that learned to predict CR and CCS, respectively. 
Each of CR-FIQA(S) (On top) and CCS-FIQA(S) (On top) is fine-tuned for 32K iteration with an initial learning rate of 0.01. 
The learning rate is divided by 10 at 20K and 28K training iterations, similarly to CR-FIAQ(S) and CCS-FIAQ(S).
The results (ERCs and AUCs) of these models are compared to CR-FIQA(S) and CCS-FIQA(S) in Figure \ref{fig:ablation}. The ERCs in Figure \ref{fig:ablation} presented the evaluation on Adience, AgeDb-30, CALFW and CFP-FP using the ArcFace FR model. 
The ERCs and AUCs in Figure \ref{fig:ablation} show that both CR-FIQA(S) and CCS-FIQA(S) lead to stronger reductions in the error than the CR-FIQA(S) (On top) and the CCS-FIQA(S) (On top), when rejecting low-quality samples.  This supports our training paradigm that simultaneously learns the internal network observation, CR, while optimizing the class centers. 
This can be related to the step-wise convergence towards the final CR value in the simultaneously training.
For both ablation study questions, ERCs and AUC for all the remaining benchmarks and FR models (mentioned in Section \ref{sec:exp}) lead to similar conclusions and are provided in the supplementary material.



\begin{figure*}[!]
\centering
\foreach \db in {Adience, AgeDB-30, CFP-FP, LFW, CALFW, CPLFW, XQLFW, IJB-C}{
    \foreach \model in {ArcFace,ElasticFace}
        {\includegraphics[scale=0.20]{images/Ercs/\db/\db_0.001_\model.pdf}}
}
\includegraphics[scale=0.33]{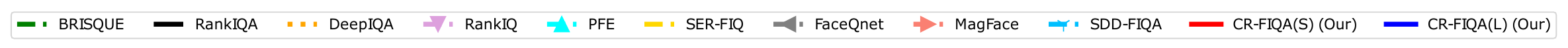}
\caption{ERC (FNMR at FMR1e-3 vs reject) curves for all evaluated benchmarks using ArcFace and ElasticFace FR models corresponding to Table \ref{tab:erc} results.  The visual evaluation, ERC curves, using MagFace and CurricularFace FR models are provided in the supplementary material. 
The proposed CR-FIQA(L) and  CR-FIQA(S) are marked with solid blue and red lines, respectively. 
CR-FIQA leads to lower verification error, when rejecting a fraction of images, of the lowest quality, in comparison to SOTA methods (faster decaying curve) under most experimental settings.
}
\label{fig:erc}
\end{figure*}

\section{Result and Discussion}
\label{sec:result}
All CR-FIQA performances reported in this paper are obtained under cross-model settings. The proposed CR-FIQA is used only to predict FIQ and not to extract feature representation of face images.
None of the utilized FR models (ArcFace \cite{deng2019arcface}, ElasticFace \cite{elasticface}, MagFace \cite{MagFace}, and CurricularFace \cite{curricularFace}) is trained with our paradigm. Instead, we trained a separate model for CR-FIQA and used the official pretrained FR models (as described in Section \ref{sec:exp}) for feature extraction. 
The verification performances as AUC at FMR1e-3 and FMR1e-4 are presented in Table \ref{tab:erc}. 
The visual verification performances as ERC curves (Figure \ref{fig:erc}) using ArcFace and ElasticFace FR models are reported at FMR1e-3. 
The ERC curves at FMR1e-4 and for MagFace and CurricularFace FR models at FMR1e-3 are provided in the supplementary material. 


The ERC curves (Figure \ref{fig:erc}) and the AUC values (Table \ref{tab:erc}) show that our proposed CR-FIQA(S) and CR-FIQA(L) outperformed the SOTA methods by significant margins in almost all settings.
Observing the results on IJB-C, Adience, CFP-FP, CALFW, and CPLFW at FMR1e-3 and FMR1e-4 (Figure \ref{fig:erc} and Table \ref{tab:erc}), our proposed CR-FIQA outperformed all SOTA methods on all the considered FR models. 
On the AgeDB-30 benchmark, our proposed CR-FIQA ranked first in five out of eight settings and second in the other three settings (Table \ref{tab:erc}).
On the LFW benchmark, our proposed CR-FIQA ranked behind the MagFace and the RankIQ. This is the only case that our models did not outperform all SOTA methods. However, it can be noticed from the ERC curves in Figure \ref{fig:erc} that none of the SOTA methods were able to achieve stable behavior (smoothly decaying curve) on LFW.
The main reason for such unstable ERC behavior on LFW is that the FR performance on LFW is nearly saturated (all models achieved above 99.80\% accuracy \cite{MagFace,deng2019arcface,elasticface,curricularFace}), leaving very few samples causing errors and thus lowering the significance of the measured FNMR. 


The XQLFW benchmark is derived from LFW to contain pairs with a maximum difference in quality. 
The XQLFW images are chosen based on BRISQUE\cite{BRISQE_IQA} and SER-FIQ \cite{XQLFW} quality scores to be either extremely high or low. 
The use of SER-FIQ \cite{SERFIQ} in this selection might give a biased edge for SER-FIQ on this benchmark.  
On XQLFW, our CR-FIQA achieved very close performance to the selection method (SER-FIQ) and is far ahead of all other SOTA methods. 
Lastly, our proposed CR-FIQA(S) achieved very comparable performance to our CR-FIQA(L), pointing out the robustness of our approach, regardless of the training database and architecture complexity. 

Table \ref{tab:ijbc_ver} presents the verification performance on the IJB-C 1:1 mixed verification benchmark \cite{ijbc} using quality scores as an embedding weighting term (as defined in \cite{ijbc}) under different experimental settings. 
For each of the FR models, we report in the first row the evaluation result of the corresponding FR model as defined in the protocol \cite{ijbc} and the corresponding released evaluation scripts \cite{deng2019arcface,elasticface,MagFace,curricularFace}, i.e. without considering the FIQ. 
Our proposed CR-FIQA significantly leads to higher verification performance than all evaluated SOTA methods, when the quality score is used as an embedding weighting term (Table \ref{tab:ijbc_ver}). 
This achievement is observable under all experimental settings (Table \ref{tab:ijbc_ver}).
Another outcome of this evaluation is that the integration of CR-FIQA leads to SOTA verification performance on one of the most challenging FR benchmarks, IJB-C \cite{ijbc}. 


\section{Conclusion}
In this work, we propose the CR-FIQA approach that probes the relative classifiability of training samples of the FR model and utilize this observation to learn to predict the utility of any given sample in achieving an accurate FR performance.
We experimentally prove the theorized relationship between the sample relative classifiability and FIQ and build on that towards our CR-FIQA. 
The CR-FIQA training paradigm simultaneously learns to optimize the class center while learning to predict sample relative classifiability.
The presented ablation studies and the extensive experimental results prove the effectiveness of the proposed CR-FIQA approach, and its design choices, as an FIQ method.
The reported results demonstrated that our proposed CR-FIQA outperformed SOTA methods repeatedly across multiple FR models and on multiple benchmarks, including ones with a large age gap (AgeDb-30, Adience, CALFW), large quality difference   (XQLFW), large pose variation (CPLFW, CFP-FP), and extremely large-scale and challenging FR benchmarks (IJB-C).

\textbf{Acknowledgment}
This research work has been funded by the German Federal Ministry of Education and Research and the Hessian Ministry of Higher Education, Research, Science and the Arts within their joint support of the National Research Center for Applied Cybersecurity ATHENE. This work has been partially funded by the German Federal Ministry of Education and Research (BMBF) through the Software Campus Project.

{\small
\bibliographystyle{ieee_fullname}
\bibliography{main}

\begin{thebibliography}{10}\itemsep=-1pt

\bibitem{bipa}
{740 ILCS/14}.
\newblock {Biometric Information Privacy Act (BIPA)}.
\newblock Public act 095-994, Illinois General Assembly, 2008.

\bibitem{best2018learning}
Lacey Best{-}Rowden and Anil~K. Jain.
\newblock Learning face image quality from human assessments.
\newblock {\em {IEEE} Trans. Inf. Forensics Secur.}, 13(12):3064--3077, 2018.

\bibitem{DEEPIQ_IQA}
Sebastian Bosse, Dominique Maniry, Klaus{-}Robert M{\"{u}}ller, Thomas Wiegand,
  and Wojciech Samek.
\newblock Deep neural networks for no-reference and full-reference image
  quality assessment.
\newblock {\em {IEEE} Trans. Image Process.}, 27(1):206--219, 2018.

\bibitem{elasticface}
Fadi Boutros, Naser Damer, Florian Kirchbuchner, and Arjan Kuijper.
\newblock Elasticface: Elastic margin loss for deep face recognition.
\newblock {\em CoRR}, abs/2109.09416, 2021.

\bibitem{RANKIQ_FIQA}
Jiansheng Chen, Yu Deng, Gaocheng Bai, and Guangda Su.
\newblock Face image quality assessment based on learning to rank.
\newblock {\em {IEEE} Signal Process. Lett.}, 22(1):90--94, 2015.

\bibitem{deng2019arcface}
Jiankang Deng, Jia Guo, Niannan Xue, and Stefanos Zafeiriou.
\newblock Arcface: Additive angular margin loss for deep face recognition.
\newblock In {\em {IEEE} Conference on Computer Vision and Pattern Recognition,
  {CVPR} 2019, Long Beach, CA, USA, June 16-20, 2019}, pages 4690--4699.
  Computer Vision Foundation / {IEEE}, 2019.

\bibitem{eadhaar}
{e-Aadhaar - Unique Identification Authority of India}.
\newblock \url{https://eaadhaar.uidai.gov.in/}, 2015.

\bibitem{Adience}
Eran Eidinger, Roee Enbar, and Tal Hassner.
\newblock Age and gender estimation of unfiltered faces.
\newblock {\em {IEEE} Trans. Inf. Forensics Secur.}, 9(12):2170--2179, 2014.

\bibitem{frontex2015best}
Frontex.
\newblock Best practice technical guidelines for automated border control (abc)
  systems, 2015.

\bibitem{BiyingWACV}
Biying Fu, Cong Chen, Olaf Henniger, and Naser Damer.
\newblock A deep insight into measuring face image utility with general and
  face-specific image quality metrics.
\newblock {\em CoRR}, abs/2110.11111, 2021.

\bibitem{smoothl1}
Ross~B. Girshick.
\newblock Fast {R-CNN}.
\newblock In {\em 2015 {IEEE} International Conference on Computer Vision,
  {ICCV} 2015, Santiago, Chile, December 7-13, 2015}, pages 1440--1448. {IEEE}
  Computer Society, 2015.

\bibitem{NISTQuaity}
P. Grother, M.~Ngan A.~Hom, and K. Hanaoka.
\newblock Ongoing face recognition vendor test (frvt) part 5: Face image
  quality assessment (4th draft).
\newblock In {\em National Institute of Standards and Technology}. Tech. Rep.,
  Sep. 2021.

\bibitem{GT07}
P. Grother and E. Tabassi.
\newblock Performance of biometric quality measures.
\newblock {\em IEEE Trans.~on Pattern Analysis and Machine Intelligence},
  29(4):531--543, Apr. 2007.

\bibitem{guo2016ms}
Yandong Guo, Lei Zhang, Yuxiao Hu, Xiaodong He, and Jianfeng Gao.
\newblock Ms-celeb-1m: {A} dataset and benchmark for large-scale face
  recognition.
\newblock In Bastian Leibe, Jiri Matas, Nicu Sebe, and Max Welling, editors,
  {\em Computer Vision - {ECCV} 2016 - 14th European Conference, Amsterdam, The
  Netherlands, October 11-14, 2016, Proceedings, Part {III}}, volume 9907 of
  {\em Lecture Notes in Computer Science}, pages 87--102. Springer, 2016.

\bibitem{DBLP:conf/cvpr/HeZRS16}
Kaiming He, Xiangyu Zhang, Shaoqing Ren, and Jian Sun.
\newblock Deep residual learning for image recognition.
\newblock In {\em 2016 {IEEE} Conference on Computer Vision and Pattern
  Recognition, {CVPR} 2016, Las Vegas, NV, USA, June 27-30, 2016}, pages
  770--778. {IEEE} Computer Society, 2016.

\bibitem{faceqnetv1}
Javier Hernandez{-}Ortega, Javier Galbally, Julian Fi{\'{e}}rrez, and Laurent
  Beslay.
\newblock Biometric quality: Review and application to face recognition with
  faceqnet.
\newblock {\em CoRR}, abs/2006.03298, 2020.

\bibitem{hernandez2019faceqnet}
Javier Hernandez{-}Ortega, Javier Galbally, Julian Fi{\'{e}}rrez, Rudolf
  Haraksim, and Laurent Beslay.
\newblock Faceqnet: Quality assessment for face recognition based on deep
  learning.
\newblock In {\em 2019 International Conference on Biometrics, {ICB} 2019,
  Crete, Greece, June 4-7, 2019}, pages 1--8. {IEEE}, 2019.

\bibitem{LFWTech}
Gary~B. Huang, Manu Ramesh, Tamara Berg, and Erik Learned-Miller.
\newblock Labeled faces in the wild: A database for studying face recognition
  in unconstrained environments.
\newblock Technical Report 07-49, University of Massachusetts, Amherst, October
  2007.

\bibitem{curricularFace}
Yuge Huang, Yuhan Wang, Ying Tai, Xiaoming Liu, Pengcheng Shen, Shaoxin Li,
  Jilin Li, and Feiyue Huang.
\newblock Curricularface: Adaptive curriculum learning loss for deep face
  recognition.
\newblock In {\em 2020 {IEEE/CVF} Conference on Computer Vision and Pattern
  Recognition, {CVPR} 2020, Seattle, WA, USA, June 13-19, 2020}, pages
  5900--5909. Computer Vision Foundation / {IEEE}, 2020.

\bibitem{ICAO18}
{ISO/IEC JTC1 SC17 WG3}.
\newblock {Portrait Quality - Reference Facial Images for MRTD}.
\newblock International Civil Aviation Organization, 2018.

\bibitem{ISOIEC29794-1}
{ISO/IEC JTC1 SC37 Biometrics}.
\newblock {ISO/IEC 29794-1:2016 Information technology - Biometric sample
  quality - Part 1: Framework}.
\newblock International Organization for Standardization, 2016.

\bibitem{ISOIEC2382-37}
{ISO/IEC JTC1 SC37 Biometrics}.
\newblock {ISO/IEC 2382-37:2017 Information technology - Vocabulary - Part 37:
  Biometrics}.
\newblock International Organization for Standardization, 2017.

\bibitem{iso_metric}
{ISO/IEC JTC1 SC37 Biometrics}.
\newblock {ISO/IEC 19795-1:2021 Information technology — Biometric
  performance testing and reporting — Part 1: Principles and framework}.
\newblock International Organization for Standardization, 2021.

\bibitem{learntorank}
Thorsten Joachims.
\newblock Optimizing search engines using clickthrough data.
\newblock In {\em Proceedings of the Eighth {ACM} {SIGKDD} International
  Conference on Knowledge Discovery and Data Mining, July 23-26, 2002,
  Edmonton, Alberta, Canada}, pages 133--142. {ACM}, 2002.

\bibitem{XQLFW}
Martin Knoche, Stefan H{\"{o}}rmann, and Gerhard Rigoll.
\newblock Cross-quality {LFW:} {A} database for analyzing cross-resolution
  image face recognition in unconstrained environments.
\newblock {\em CoRR}, abs/2108.10290, 2021.

\bibitem{DBLP:conf/cvpr/LiuWYLRS17}
Weiyang Liu, Yandong Wen, Zhiding Yu, Ming Li, Bhiksha Raj, and Le Song.
\newblock Sphereface: Deep hypersphere embedding for face recognition.
\newblock In {\em 2017 {IEEE} Conference on Computer Vision and Pattern
  Recognition, {CVPR} 2017, Honolulu, HI, USA, July 21-26, 2017}, pages
  6738--6746. {IEEE} Computer Society, 2017.

\bibitem{liu2017sphereface}
Weiyang Liu, Yandong Wen, Zhiding Yu, Ming Li, Bhiksha Raj, and Le Song.
\newblock {SphereFace}: Deep hypersphere embedding for face recognition.
\newblock In {\em Proc.~of the IEEE Conf.~on Computer Vision and Pattern
  Recognition}, pages 212--220, 2017.

\bibitem{liu2017rankiqa}
Xialei Liu, Joost van~de Weijer, and Andrew~D. Bagdanov.
\newblock Rankiqa: Learning from rankings for no-reference image quality
  assessment.
\newblock In {\em {IEEE} International Conference on Computer Vision, {ICCV}
  2017, Venice, Italy, October 22-29, 2017}, pages 1040--1049. {IEEE} Computer
  Society, 2017.

\bibitem{ijbc}
Brianna Maze, Jocelyn~C. Adams, James~A. Duncan, Nathan~D. Kalka, Tim Miller,
  Charles Otto, Anil~K. Jain, W.~Tyler Niggel, Janet Anderson, Jordan Cheney,
  and Patrick Grother.
\newblock {IARPA} janus benchmark - {C:} face dataset and protocol.
\newblock In {\em 2018 International Conference on Biometrics, {ICB} 2018, Gold
  Coast, Australia, February 20-23, 2018}, pages 158--165. {IEEE}, 2018.

\bibitem{DBLP:journals/tifs/MedenRTDKSRPS21}
Blaz Meden, Peter Rot, Philipp Terh{\"{o}}rst, Naser Damer, Arjan Kuijper,
  Walter~J. Scheirer, Arun Ross, Peter Peer, and Vitomir Struc.
\newblock Privacy-enhancing face biometrics: {A} comprehensive survey.
\newblock {\em {IEEE} Trans. Inf. Forensics Secur.}, 16:4147--4183, 2021.

\bibitem{MagFace}
Qiang Meng, Shichao Zhao, Zhida Huang, and Feng Zhou.
\newblock Magface: {A} universal representation for face recognition and
  quality assessment.
\newblock In {\em {IEEE} Conference on Computer Vision and Pattern Recognition,
  {CVPR} 2021, virtual, June 19-25, 2021}, pages 14225--14234. Computer Vision
  Foundation / {IEEE}, 2021.

\bibitem{BRISQE_IQA}
Anish Mittal, Anush~Krishna Moorthy, and Alan~Conrad Bovik.
\newblock No-reference image quality assessment in the spatial domain.
\newblock {\em {IEEE} Trans. Image Process.}, 21(12):4695--4708, 2012.

\bibitem{nique}
Anish Mittal, Rajiv Soundararajan, and Alan~C. Bovik.
\newblock Making a "completely blind" image quality analyzer.
\newblock {\em {IEEE} Signal Process. Lett.}, 20(3):209--212, 2013.

\bibitem{agedb}
Stylianos Moschoglou, Athanasios Papaioannou, Christos Sagonas, Jiankang Deng,
  Irene Kotsia, and Stefanos Zafeiriou.
\newblock Agedb: The first manually collected, in-the-wild age database.
\newblock In {\em 2017 {IEEE} CVPRW, {CVPR} Workshops 2017, Honolulu, HI, USA,
  July 21-26, 2017}, pages 1997--2005. {IEEE} Computer Society, 2017.

\bibitem{SDDFIQA}
Fu{-}Zhao Ou, Xingyu Chen, Ruixin Zhang, Yuge Huang, Shaoxin Li, Jilin Li, Yong
  Li, Liujuan Cao, and Yuan{-}Gen Wang.
\newblock {SDD-FIQA:} unsupervised face image quality assessment with
  similarity distribution distance.
\newblock In {\em {IEEE} Conference on Computer Vision and Pattern Recognition,
  {CVPR} 2021, virtual, June 19-25, 2021}, pages 7670--7679. Computer Vision
  Foundation / {IEEE}, 2021.

\bibitem{cfp-fp}
Soumyadip Sengupta, Jun{-}Cheng Chen, Carlos~Domingo Castillo, Vishal~M. Patel,
  Rama Chellappa, and David~W. Jacobs.
\newblock Frontal to profile face verification in the wild.
\newblock In {\em 2016 {IEEE} Winter Conference on Applications of Computer
  Vision, {WACV} 2016, Lake Placid, NY, USA, March 7-10, 2016}, pages 1--9.
  {IEEE} Computer Society, 2016.

\bibitem{PFE_FIQA}
Yichun Shi and Anil~K. Jain.
\newblock Probabilistic face embeddings.
\newblock In {\em 2019 {IEEE/CVF} International Conference on Computer Vision,
  {ICCV} 2019, Seoul, Korea (South), October 27 - November 2, 2019}, pages
  6901--6910. {IEEE}, 2019.

\bibitem{SERFIQ}
Philipp Terh{\"{o}}rst, Jan~Niklas Kolf, Naser Damer, Florian Kirchbuchner, and
  Arjan Kuijper.
\newblock {SER-FIQ:} unsupervised estimation of face image quality based on
  stochastic embedding robustness.
\newblock In {\em 2020 {IEEE/CVF} Conference on Computer Vision and Pattern
  Recognition, {CVPR} 2020, Seattle, WA, USA, June 13-19, 2020}, pages
  5650--5659. Computer Vision Foundation / {IEEE}, 2020.

\bibitem{GDPR_practice}
Paul Voigt and Axel von~dem Bussche.
\newblock {\em The EU General Data Protection Regulation (GDPR): A Practical
  Guide}.
\newblock 1st edition, 2017.

\bibitem{DBLP:conf/cvpr/WangWZJGZL018}
Hao Wang, Yitong Wang, Zheng Zhou, Xing Ji, Dihong Gong, Jingchao Zhou, Zhifeng
  Li, and Wei Liu.
\newblock Cosface: Large margin cosine loss for deep face recognition.
\newblock In {\em 2018 {IEEE} Conference on Computer Vision and Pattern
  Recognition, {CVPR} 2018, Salt Lake City, UT, USA, June 18-22, 2018}, pages
  5265--5274. {IEEE} Computer Society, 2018.

\bibitem{ijbb}
Cameron Whitelam, Emma Taborsky, Austin Blanton, Brianna Maze, Jocelyn~C.
  Adams, Tim Miller, Nathan~D. Kalka, Anil~K. Jain, James~A. Duncan, Kristen
  Allen, Jordan Cheney, and Patrick Grother.
\newblock {IARPA} janus benchmark-b face dataset.
\newblock In {\em 2017 {IEEE} Conference on Computer Vision and Pattern
  Recognition Workshops, {CVPR} Workshops 2017, Honolulu, HI, USA, July 21-26,
  2017}, pages 592--600. {IEEE} Computer Society, 2017.

\bibitem{pcnet}
Weidi Xie, Jeffrey Byrne, and Andrew Zisserman.
\newblock Inducing predictive uncertainty estimation for face verification.
\newblock In {\em 31st British Machine Vision Conference 2020, {BMVC} 2020,
  Virtual Event, UK, September 7-10, 2020}. {BMVA} Press, 2020.

\bibitem{casia_webface}
Dong Yi, Zhen Lei, Shengcai Liao, and Stan~Z. Li.
\newblock Learning face representation from scratch.
\newblock {\em CoRR}, abs/1411.7923, 2014.

\bibitem{CPLFWTech}
T. Zheng and W. Deng.
\newblock Cross-pose lfw: A database for studying cross-pose face recognition
  in unconstrained environments.
\newblock Technical Report 18-01, Beijing University of Posts and
  Telecommunications, February 2018.

\bibitem{CALFW}
Tianyue Zheng, Weihong Deng, and Jiani Hu.
\newblock Cross-age {LFW:} {A} database for studying cross-age face recognition
  in unconstrained environments.
\newblock {\em CoRR}, abs/1708.08197, 2017.

\end{thebibliography}
}

\section{Supplementary Material}
This supplementary material complements the main submission by providing:
\begin{enumerate}
    \item Complementary ERC curves with AUC values for all the FR models and benchmarks to complement and support the ablation study section (Section5) of the main manuscript.
    \item Samples images from the 8 benchmarks with quality scores achieved by our CR-FIQA and SOTA methods.
    \item Quality score distribution of the evaluation benchmarks achieved by our CR-FIQA and SOTA methods.
    \item  ERC (FNMR at FMR1e-4 vs reject) curves that provide a complement to the AUC reported in Table 1 of the main manuscript.
    \item  ERC (FNMR at FMR1e-3 vs reject) curves using MagFace and Curricular FR models that provide a complement to the AUC reported in Table 1 and Figure 4 of the main manuscript.
    \item More details on the databases and benchmarks. 
    \item  Discussion of the potential social impacts.
    \item Details on further existing assets used in the work.
    \item A discussion on the technical limitations of the presented work.
\end{enumerate}

\subsection{Complementary Result for Ablation Study}
\begin{figure*}[htb]
\centering
\foreach \db in {Adience, AgeDb-30, CFP-FP}
{
    \foreach \model in {ArcFaceModel, ElasticFaceModel}
    {\includegraphics[scale=0.40]{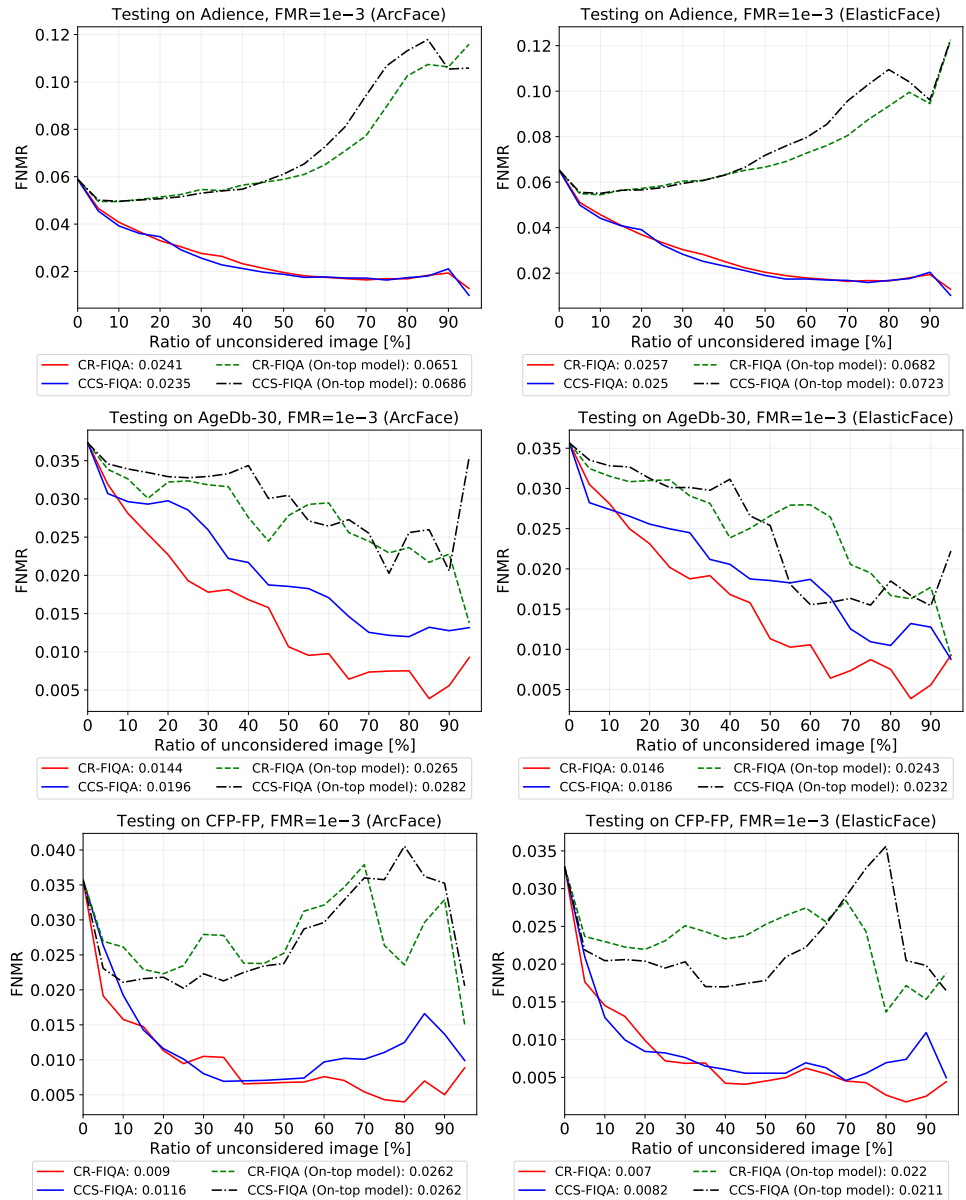}}
}
\caption{ERC comparison between CR-FIQA(S), CCS-FIQA(S), CR-FIQA(S) (On top) and CCS-FIQA(S) (On top). 
The plots show the effect of rejecting samples of lowest quality, on the verification error (FNMR at FMR1e-3) using ArcFace and ElasticFace models on Adience, AgeDb-30 and CFP-FP benchmarks . CR-FIQA(S) and CCS-FIQA(S) outperformed the on-top solutions, and CR-FIQA(S) performs generally better than CCS-FIQA(S) (curve decays faster with more rejected samples). AUC values are mentioned under the plots. }
\label{fig:ablation_study_3_1}
\end{figure*}

\begin{figure*}[htb]
\centering
\foreach \db in {LFW, CALFW, CPLFW, XQLFW}
{
    \foreach \model in {ArcFaceModel, ElasticFaceModel}
    {\includegraphics[scale=0.40]{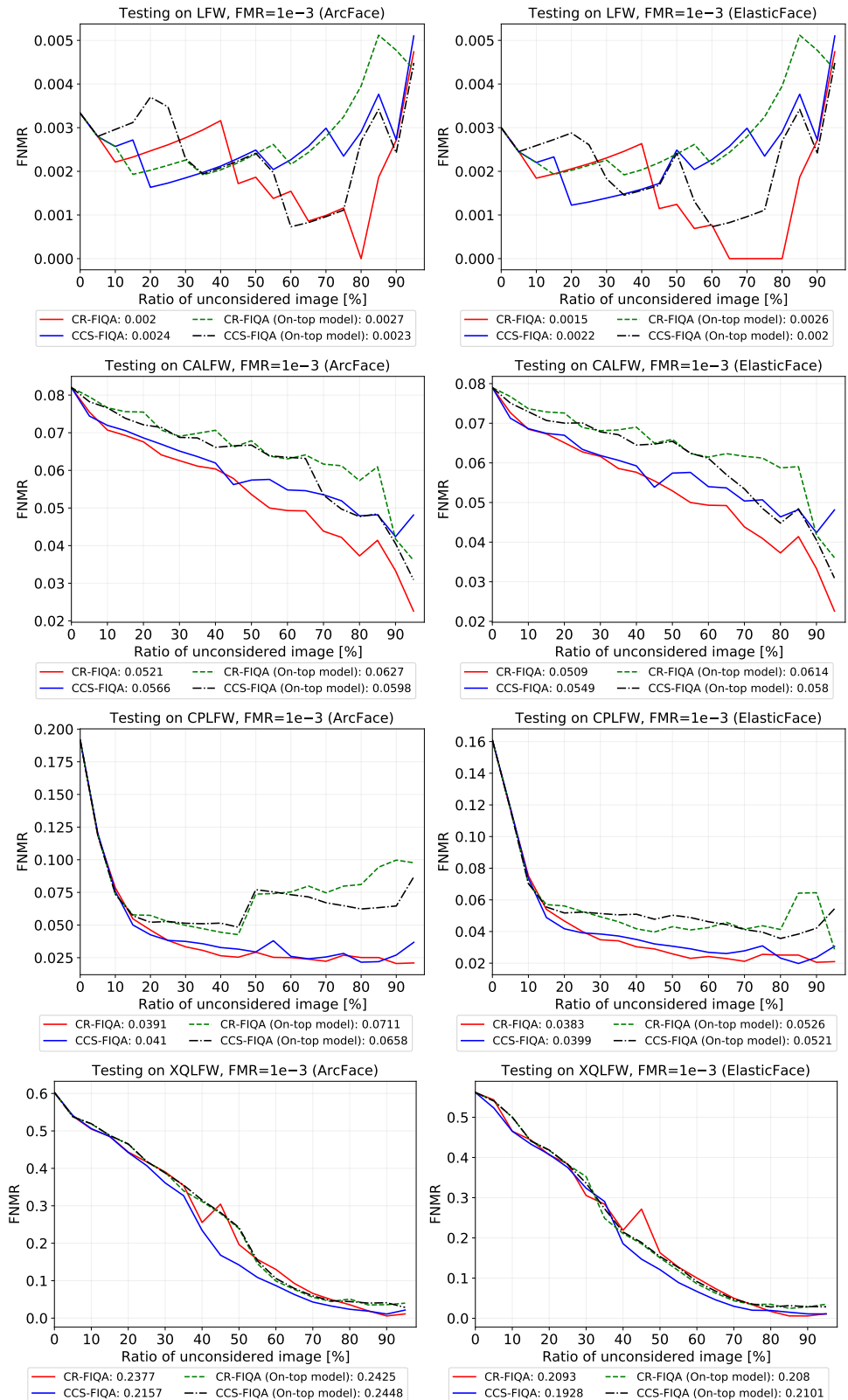}}
}
\vspace{-3mm}
\caption{ERC comparison between CR-FIQA(S), CCS-FIQA(S), CR-FIQA(S) (On top) and CCS-FIQA(S) (On top). 
The plots show the effect of rejecting samples of lowest quality, on the verification error (FNMR at FMR1e-3) using ArcFace and ElasticFace models on LFW, CALFW, CPLFW and XQLFW benchmarks . CR-FIQA(S) and CCS-FIQA(S) outperformed the on-top solutions, and CR-FIQA(S) performs generally better than CCS-FIQA(S) (curve decays faster with more rejected samples). AUC values are mentioned under the plots. }
\label{fig:ablation_study_3_2}
\end{figure*}

\begin{figure*}[htb]
\centering
\foreach \db in {Adience, AgeDb-30, CFP-FP}
{
    \foreach \model in {MagFaceModel, CurricularFaceModel}
    {\includegraphics[scale=0.40]{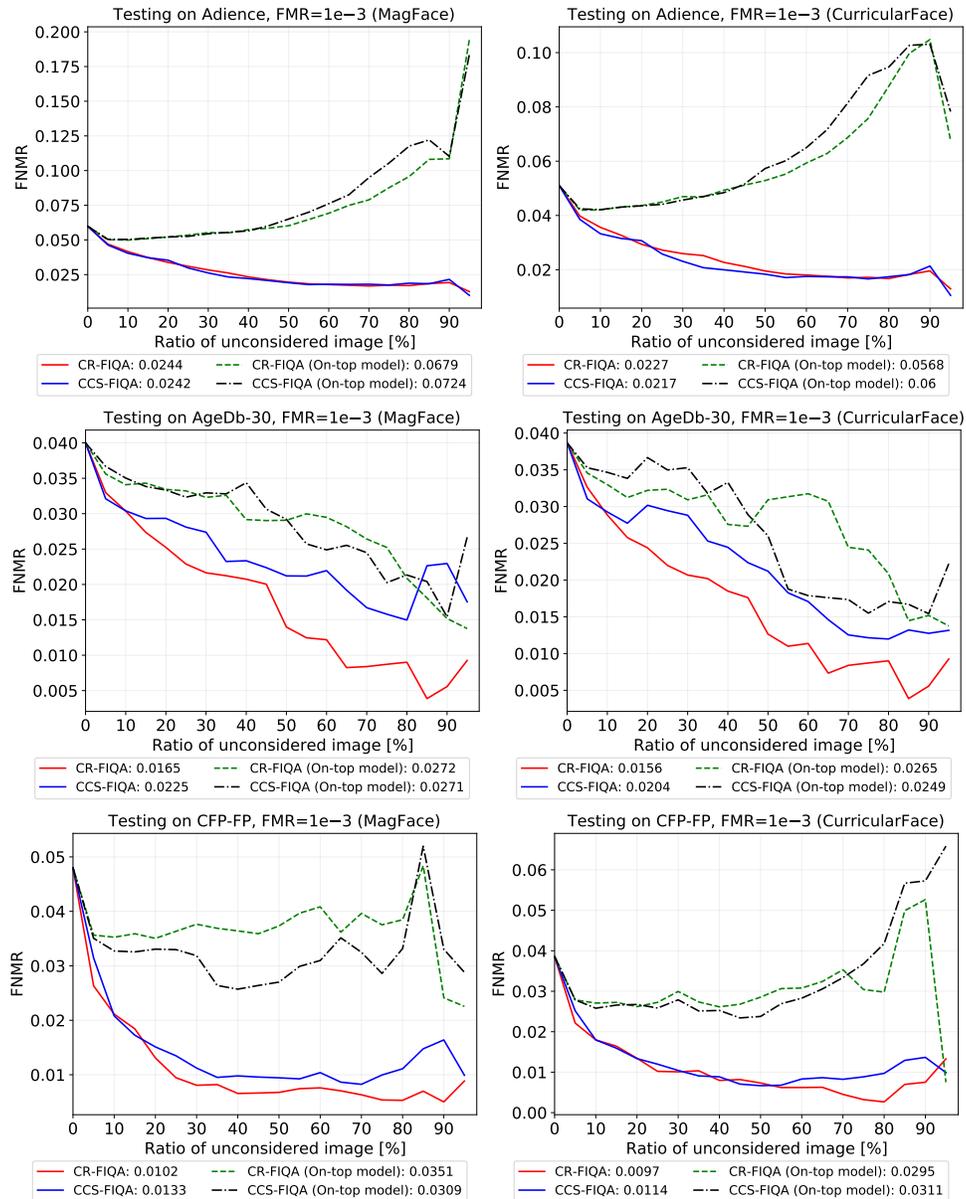}}
}
\caption{ERC comparison between CR-FIQA(S), CCS-FIQA(S), CR-FIQA(S) (On top) and CCS-FIQA(S) (On top). 
The plots show the effect of rejecting samples of lowest quality, on the verification error (FNMR at FMR1e-3) using MagFace and CurricularFace models on Adience, AgeDb-30 and CFP-FP  benchmarks . CR-FIQA(S) and CCS-FIQA(S) outperformed the on-top solutions, and CR-FIQA(S) performs generally better than CCS-FIQA(S) (curve decays faster with more rejected samples). AUC values are mentioned under the plots. }
\label{fig:ablation_study_3_3}
\end{figure*}

\begin{figure*}[htb]
\centering
\foreach \db in { LFW, CALFW, CPLFW, XQLFW}
{
    \foreach \model in {MagFaceModel, CurricularFaceModel}
    {\includegraphics[scale=0.40]{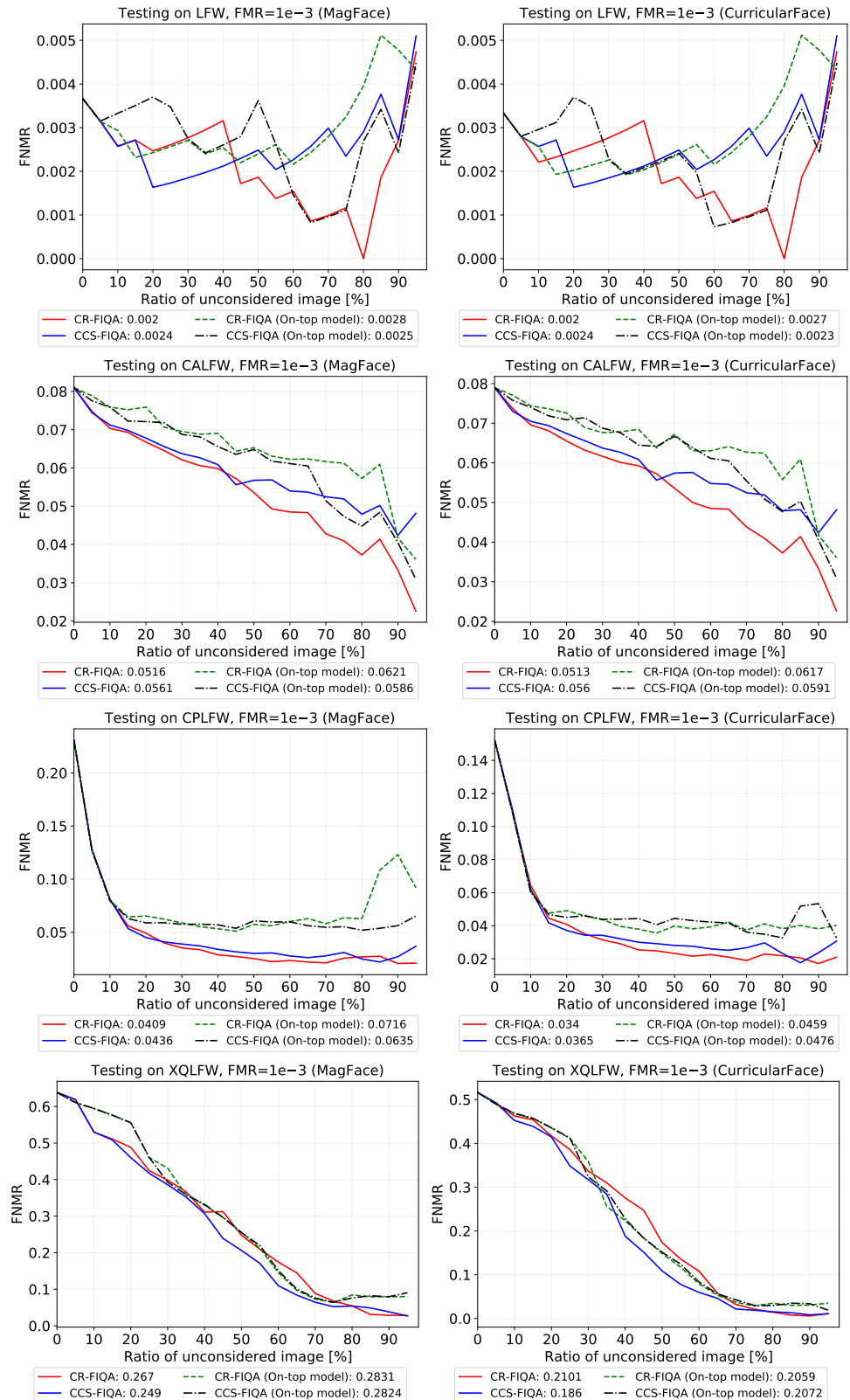}}
}
\vspace{-3mm}
\caption{ERC comparison between CR-FIQA(S), CCS-FIQA(S), CR-FIQA(S) (On top) and CCS-FIQA(S) (On top). 
The plots show the effect of rejecting samples of lowest quality, on the verification error (FNMR at FMR1e-3) using MagFace and CurricularFace models on LFW, CALFW, CPLFW and XQLFW benchmarks . CR-FIQA(S) and CCS-FIQA(S) outperformed the on-top solutions, and CR-FIQA(S) performs generally better than CCS-FIQA(S) (curve decays faster with more rejected samples). AUC values are mentioned under the plots. }
\label{fig:ablation_study_3_4}
\end{figure*}

\begin{figure*}[htb]
\centering
\foreach \db in {Adience, AgeDb-30, CFP-FP}
{
    \foreach \model in {ArcFaceModel, ElasticFaceModel}
    {\includegraphics[scale=0.40]{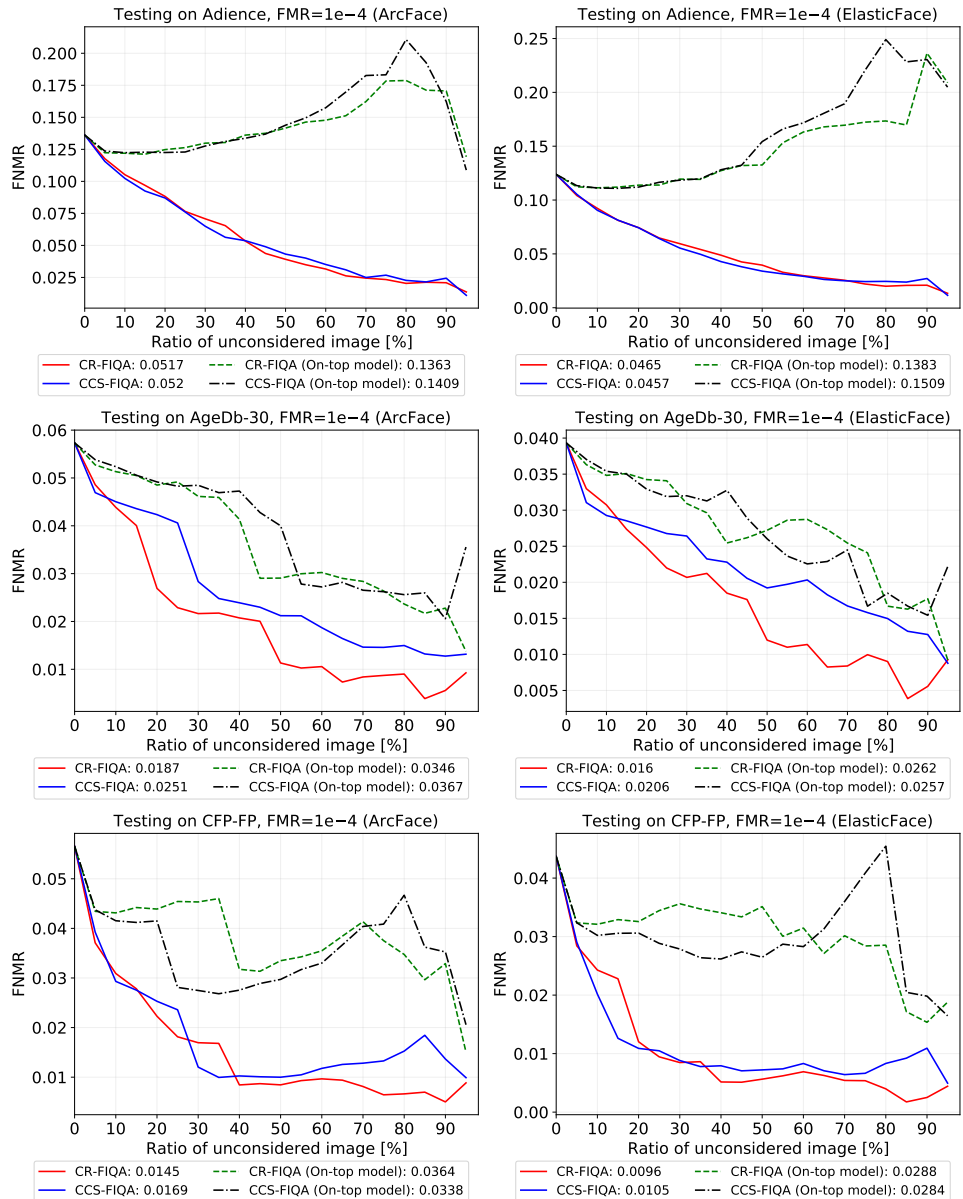}}
    
}
\caption{ERC comparison between CR-FIQA(S), CCS-FIQA(S), CR-FIQA(S) (On top) and CCS-FIQA(S) (On top). 
The plots show the effect of rejecting samples of lowest quality, on the verification error (FNMR at FMR1e-4) using ArcFace and ElasticFace models on Adience, AgeDb-30 and CFP-FP benchmarks . CR-FIQA(S) and CCS-FIQA(S) outperformed the on-top solutions, and CR-FIQA(S) performs generally better than CCS-FIQA(S) (curve decays faster with more rejected samples). AUC values are mentioned under the plots. }
\label{fig:ablation_study_4_1}
\end{figure*}

\begin{figure*}[htb]
\centering
\foreach \db in { LFW, CALFW, CPLFW, XQLFW}
{
    \foreach \model in {ArcFaceModel, ElasticFaceModel}
    {\includegraphics[scale=0.40]{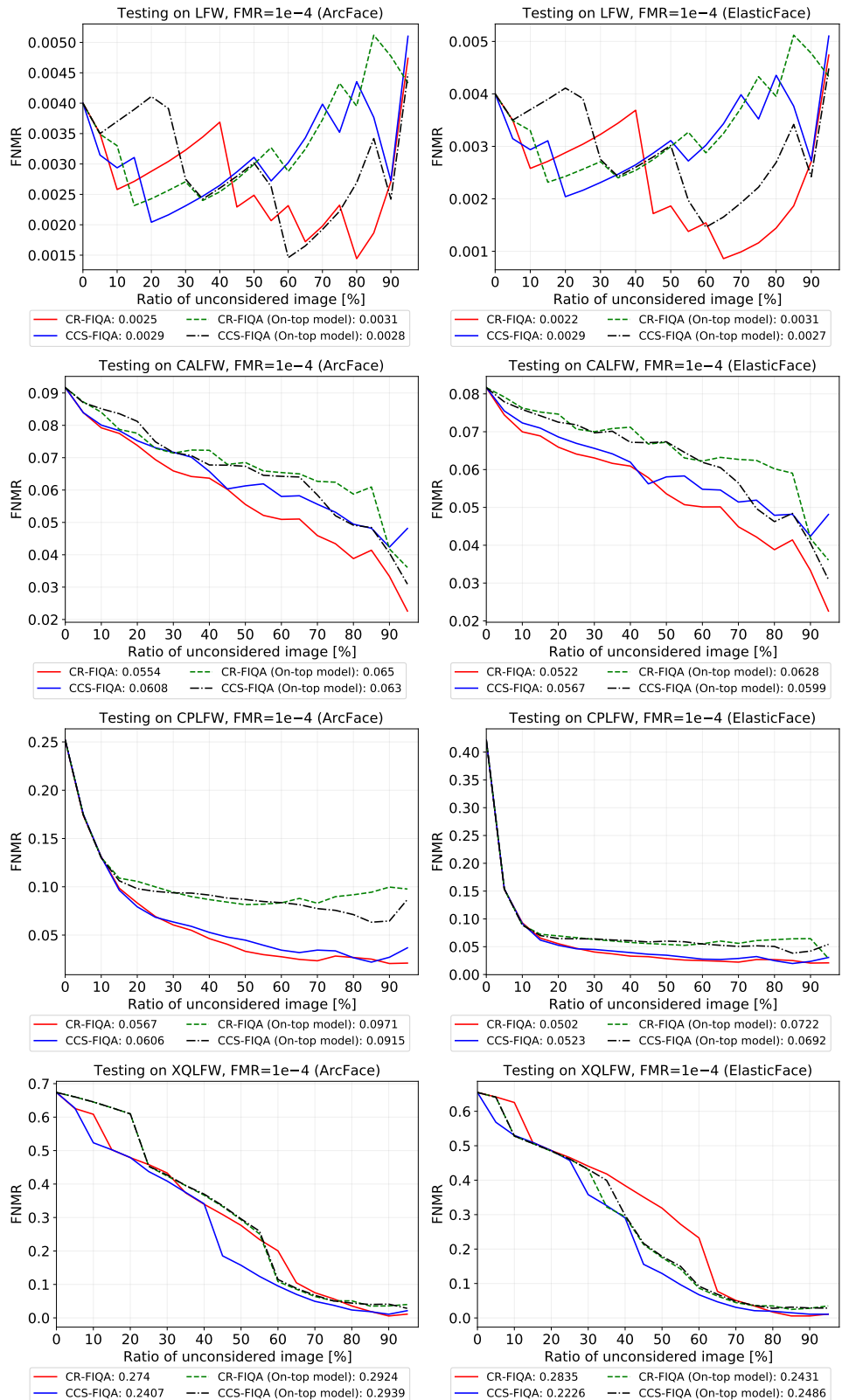}}
    
}
\vspace{-3mm}
\caption{ERC comparison between CR-FIQA(S), CCS-FIQA(S), CR-FIQA(S) (On top) and CCS-FIQA(S) (On top). 
The plots show the effect of rejecting samples of lowest quality, on the verification error (FNMR at FMR1e-4) using ArcFace and ElasticFace models on LFW, CALFW, CPLFW and XQLFW benchmarks . CR-FIQA(S) and CCS-FIQA(S) outperformed the on-top solutions, and CR-FIQA(S) performs generally better than CCS-FIQA(S) (curve decays faster with more rejected samples). AUC values are mentioned under the plots. }
\label{fig:ablation_study_4_2}
\end{figure*}

\begin{figure*}[htb]
\centering
\foreach \db in {Adience, AgeDb-30, CFP-FP}
{
    \foreach \model in {MagFaceModel, CurricularFaceModel}
    {\includegraphics[scale=0.40]{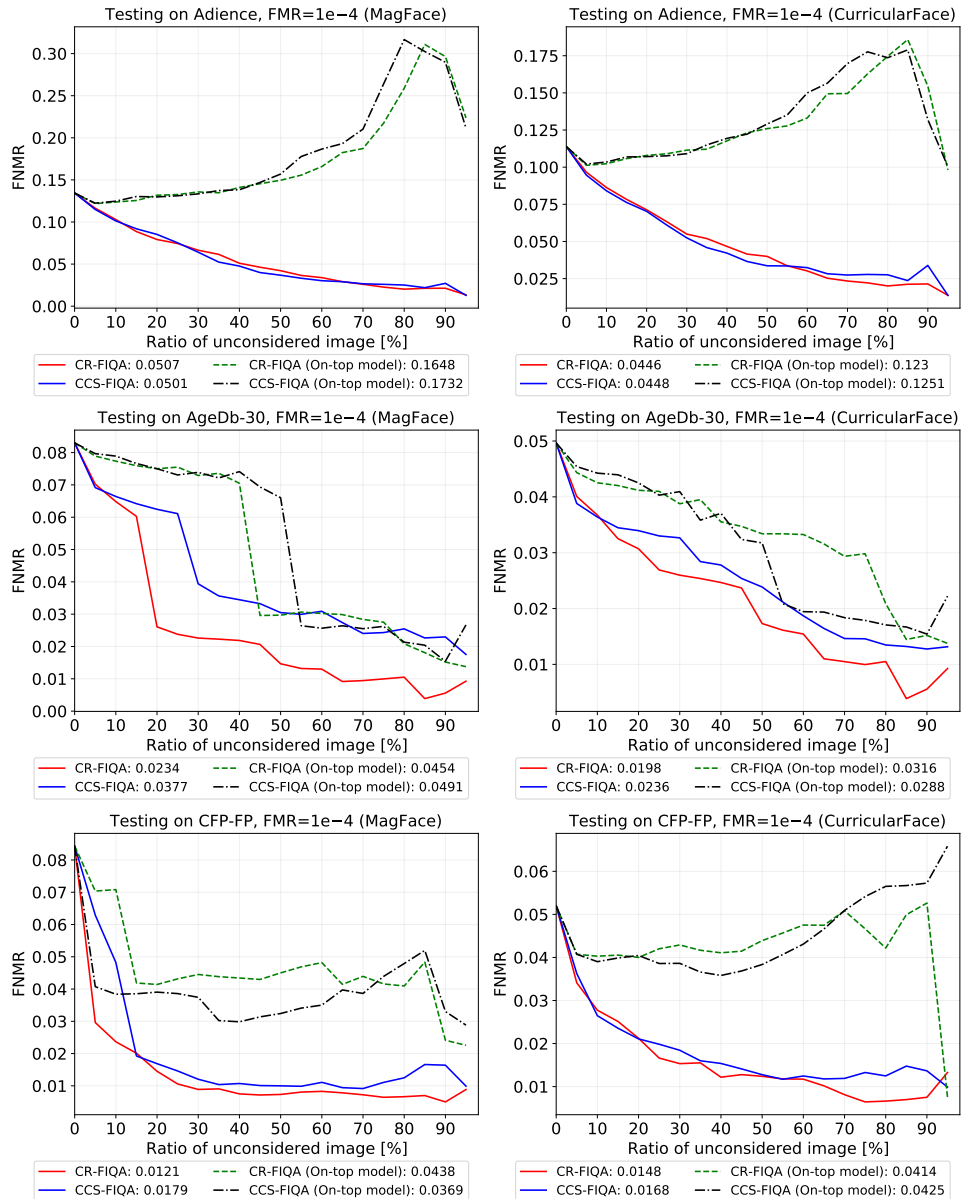}}
    
}
\caption{ERC comparison between CR-FIQA(S), CCS-FIQA(S), CR-FIQA(S) (On top) and CCS-FIQA(S) (On top). 
The plots show the effect of rejecting samples of lowest quality, on the verification error (FNMR at FMR1e-4) using MagFace and CurricularFace on Adience, AgeDb-30 and CFP-FP benchmarks . CR-FIQA(S) and CCS-FIQA(S) outperformed the on-top solutions, and CR-FIQA(S) performs generally better than CCS-FIQA(S) (curve decays faster with more rejected samples). AUC values are mentioned under the plots. }
\label{fig:ablation_study_4_3}
\end{figure*}

\begin{figure*}[htb]
\centering
\foreach \db in { LFW, CALFW, CPLFW, XQLFW}
{
    \foreach \model in {MagFaceModel, CurricularFaceModel}
    {\includegraphics[scale=0.40]{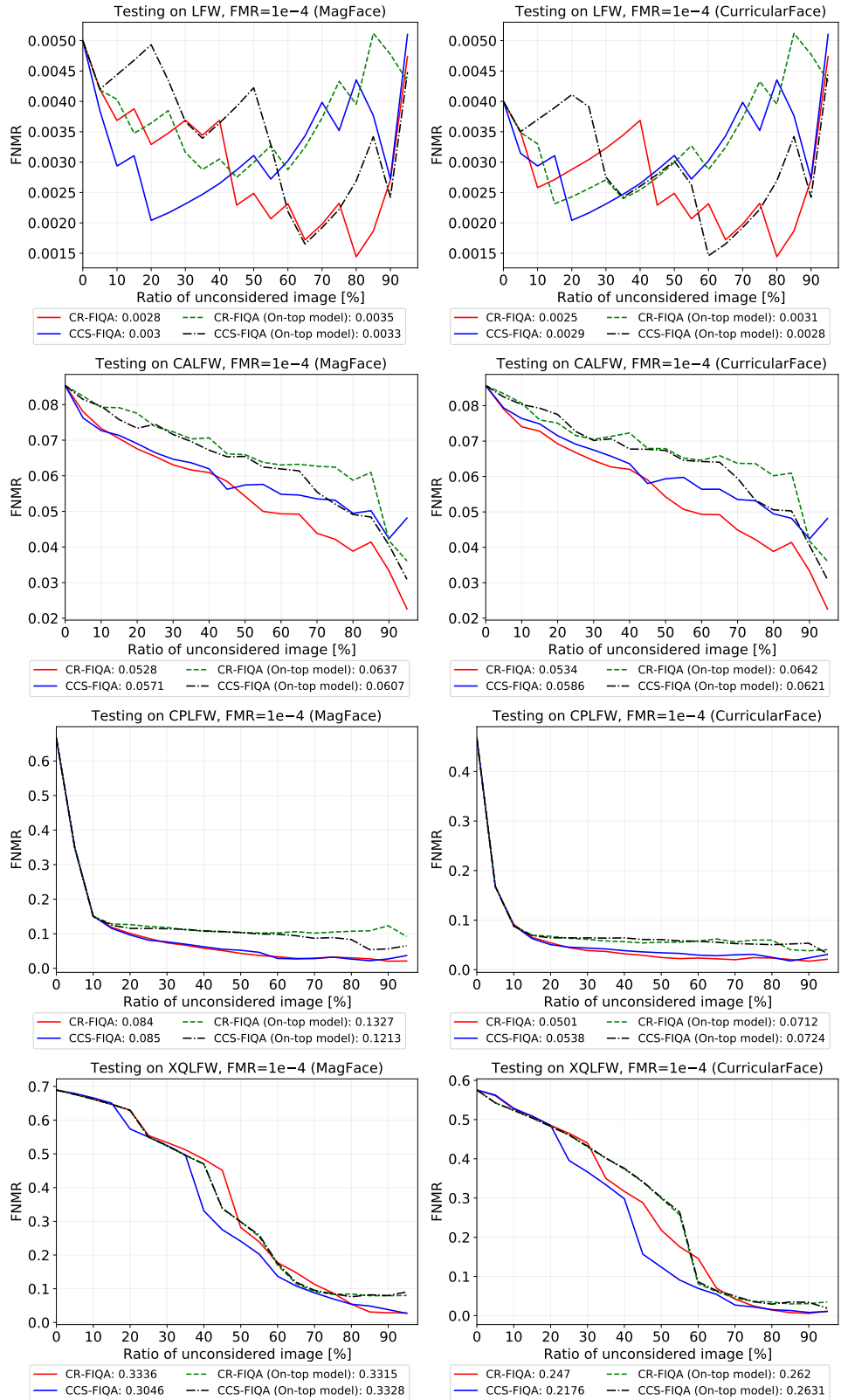}}
    
}
\vspace{-3mm}
\caption{ERC comparison between CR-FIQA(S), CCS-FIQA(S), CR-FIQA(S) (On top) and CCS-FIQA(S) (On top). 
The plots show the effect of rejecting samples of lowest quality, on the verification error (FNMR at FMR1e-4) using MagFace and CurricularFace on LFW, CALFW, CPLFW and XQLFW benchmarks . CR-FIQA(S) and CCS-FIQA(S) outperformed the on-top solutions, and CR-FIQA(S) performs generally better than CCS-FIQA(S) (curve decays faster with more rejected samples). AUC values are mentioned under the plots. }\label{fig:ablation_study_4_4}
\end{figure*}

Figures \ref{fig:ablation_study_3_1}, \ref{fig:ablation_study_3_2}, \ref{fig:ablation_study_3_3} and \ref{fig:ablation_study_3_4} present a comparison between ERCs (FNMR at FMR1e-3) of CR-FIQA(S), CCS-FIQA(S), CR-FIQA(S) (On top) and CCS-FIQA(S) (On top) on the evaluation benchmarks. 
Figures \ref{fig:ablation_study_4_1}, \ref{fig:ablation_study_4_2}, \ref{fig:ablation_study_4_3} and \ref{fig:ablation_study_4_4} present a comparison between ERCs (FNMR at FMR1e-4) of CR-FIQA(S), CCS-FIQA(S), CR-FIQA(S) (On top), and CCS-FIQA(S) (On top) on the evaluation benchmarks. 
These ERC curves are complementary to the ablation study presented in main manuscript  (Section 5).  
In the main submission, these ERC curves are presented for ArcFace \cite{deng2019arcface} FR model on Adience \cite{Adience}, AgeDb-30 \cite{agedb}, CALFW \cite{CALFW} and CFP-FP \cite{cfp-fp} in Figure 3 (main submission) and discussed on ablation study section  (Section 5 of main submission).
However, in this supplementary material we opt to provide the evaluation mentioned in Lines 583-597 on all considered FR models and evaluation benchmarks to stress the conclusion of our ablation study (Section 5 of main submission). This again points out the benefits of CR of CCS (thus the NNCCS term in equation 4 of the main submission), as well as the simultaneously training rather than on the top learning.

\subsection{Histogram of CCS and NNCCS}
\begin{figure*}[t]
  \centering
     \begin{subfigure}[b]{0.49\textwidth}
         \centering
   \includegraphics[width=\textwidth]{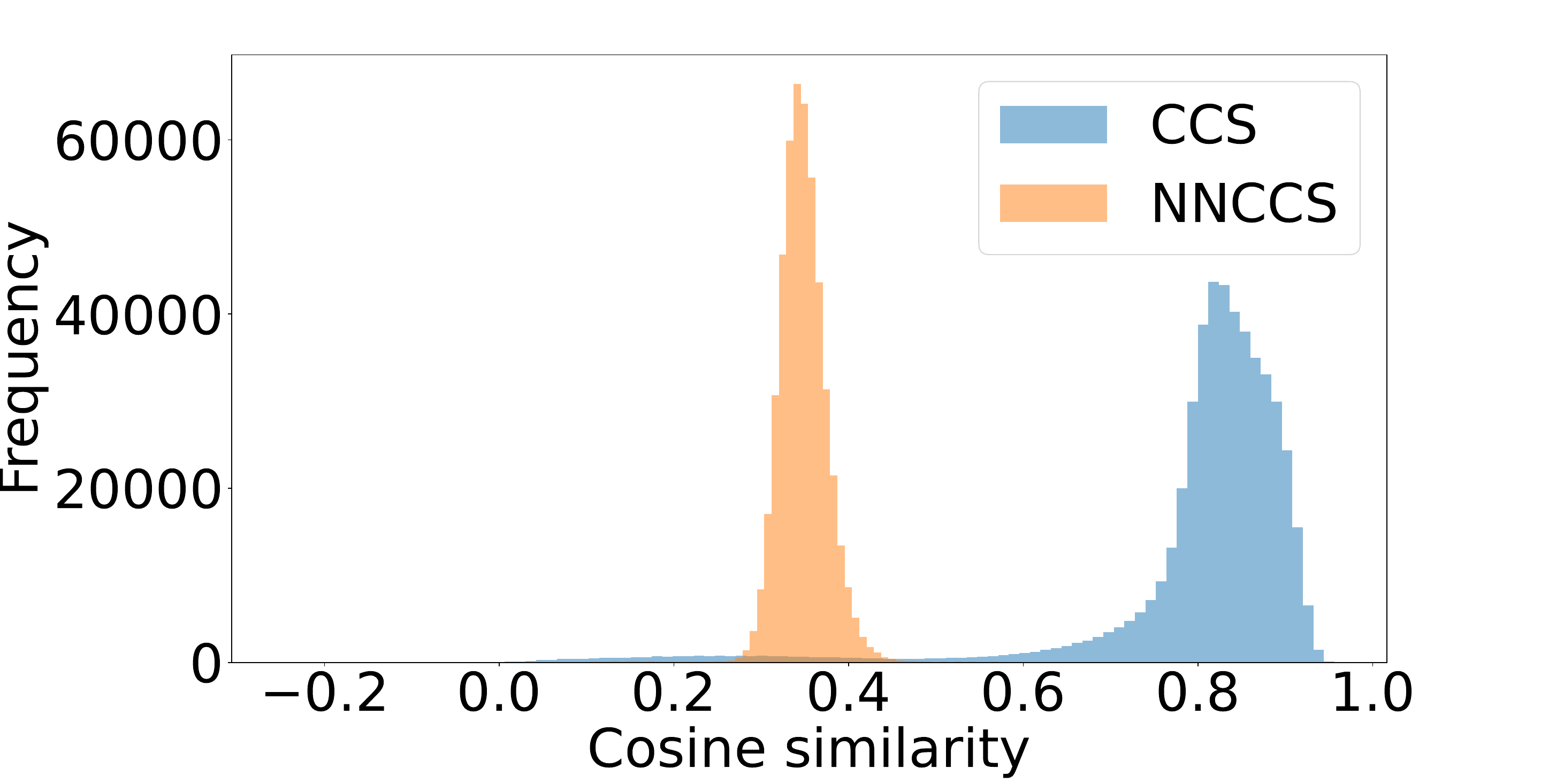}
         \caption{}
         \label{fig:histogram_casia_sp}
     \end{subfigure}
          \hfill
     \begin{subfigure}[b]{0.49\textwidth}
         \centering
   \includegraphics[width=\textwidth]{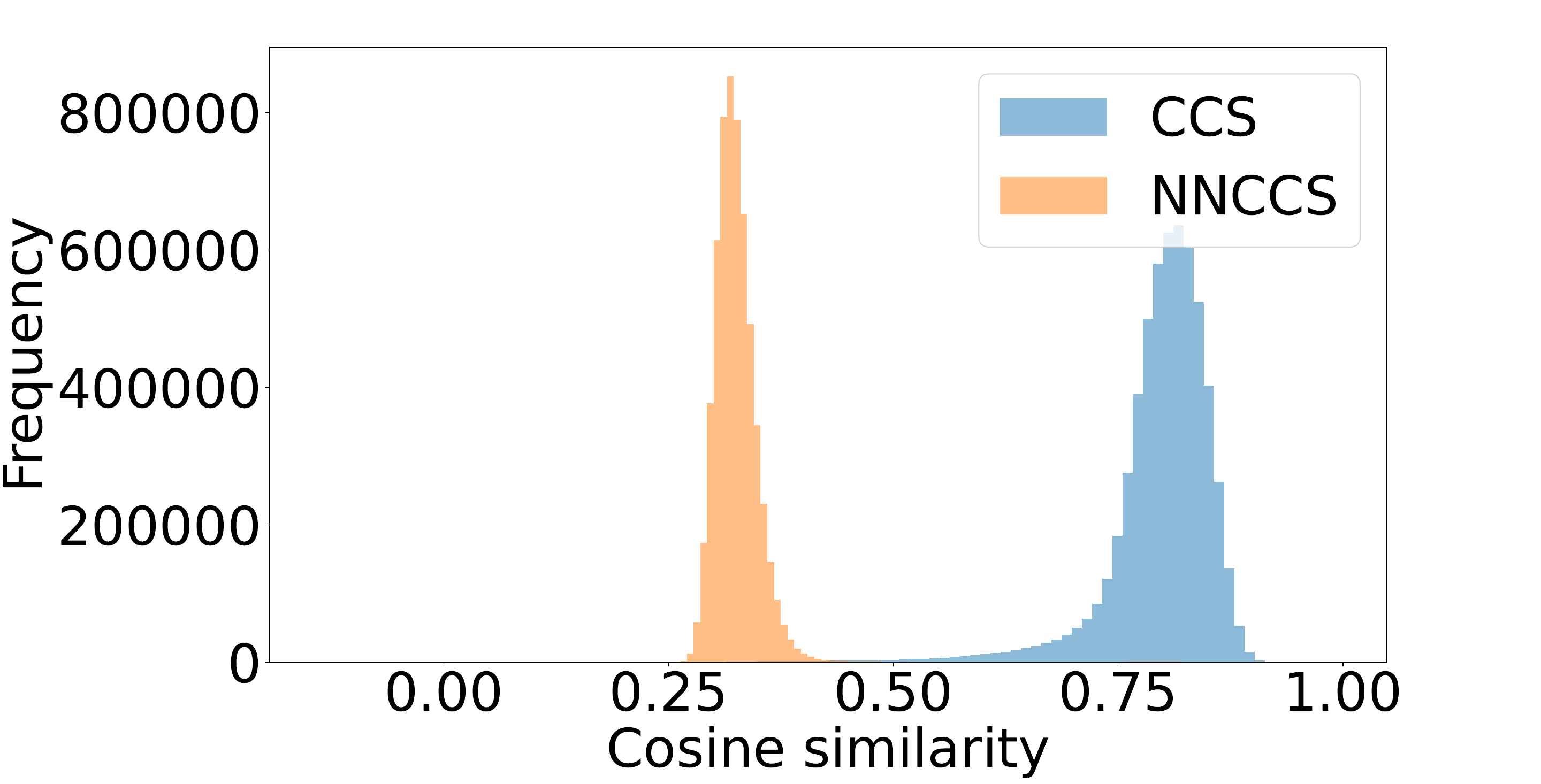}
   \caption{}
         \label{fig:histogram_ms1mv2_sp}
     \end{subfigure}
   \caption{ 
   Histogram of the cosine similarity between training samples and their class centers (CCS) and nearest negative class centers (NNCCS). Similarity values in plot \ref{fig:histogram_casia_sp} are
   obtained from ResNet-50 trained on CASIA-WebFace (R50(CASIA)) and the ones in plot \ref{fig:histogram_ms1mv2_sp} are obtained from ResNet-100 trained on MS1MV2 (R100(MS1MV2)). In both models/databases, the values of CCS and NNCCS vary between different samples.
   }
          \label{fig:histogram_distribution_sp}
\end{figure*}

Figure \ref{fig:histogram_distribution_sp} shows an insight into the CCS and NNCCS values distribution of the training datasets (CASIA-WebFace \cite{casia_webface} and MS1MV2 \cite{deng2019arcface}). Figure \ref{fig:histogram_casia_sp} shows an enhanced visualisation of the same plot shown in Figure 1 (main submission) based on the R50(CASIA) model and discussed in lines 338-342 of the main submission.
Figure \ref{fig:histogram_ms1mv2_sp} shows CCS and NNCCS values distribution of MS1MV2 dataset obtained from ResNet-100 (R100(MS1M-V2)) model to provide an additional illustration of the CCS and NNCCS value distribution on another training setup (model and dataset). On both models one can notice that the CCS and NNCCS values vary between samples.

\subsection{Quality score distribution}
Figure \ref{fig:score_distribution} presents the quality score distribution of the evaluation benchmarks achieved by our CR-FIQA and the SOTA methods, all normalized to have a range between 0 and 1. 
One can notice in the distributions, that for the XQLFW dataset where the data contains extreme low and extreme low quality samples by design, this two groups of quality is only visible in our CR-FIQA, PFE, MagFace, SDD-FIQA, as well as the methods that were used to label the qualities when constructing the XQLFW, i.e. SER-FIQA and BRISQUE.

\subsection{Sample images with quality scores}
\begin{figure*}[htb]
\begin{center}
\includegraphics[width=0.95\linewidth]{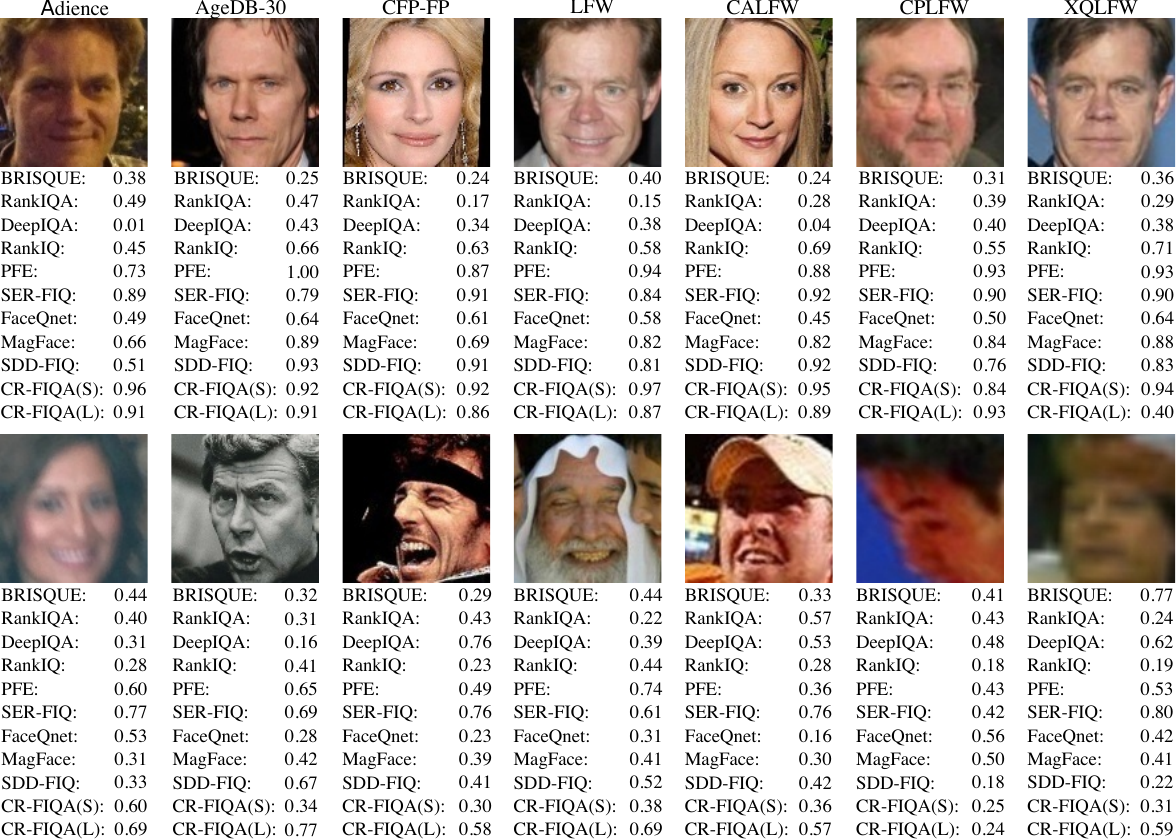}
\end{center}
\caption{Samples image of the evaluation benchmarks with quality score values obtained from our CR-FIQA  the SOTA methods. Noting that this figure only reflects samples with quality scores and do not necessary reflect overall performance.}
\label{fig:image_examples}
\end{figure*}

\begin{figure*}[htb]
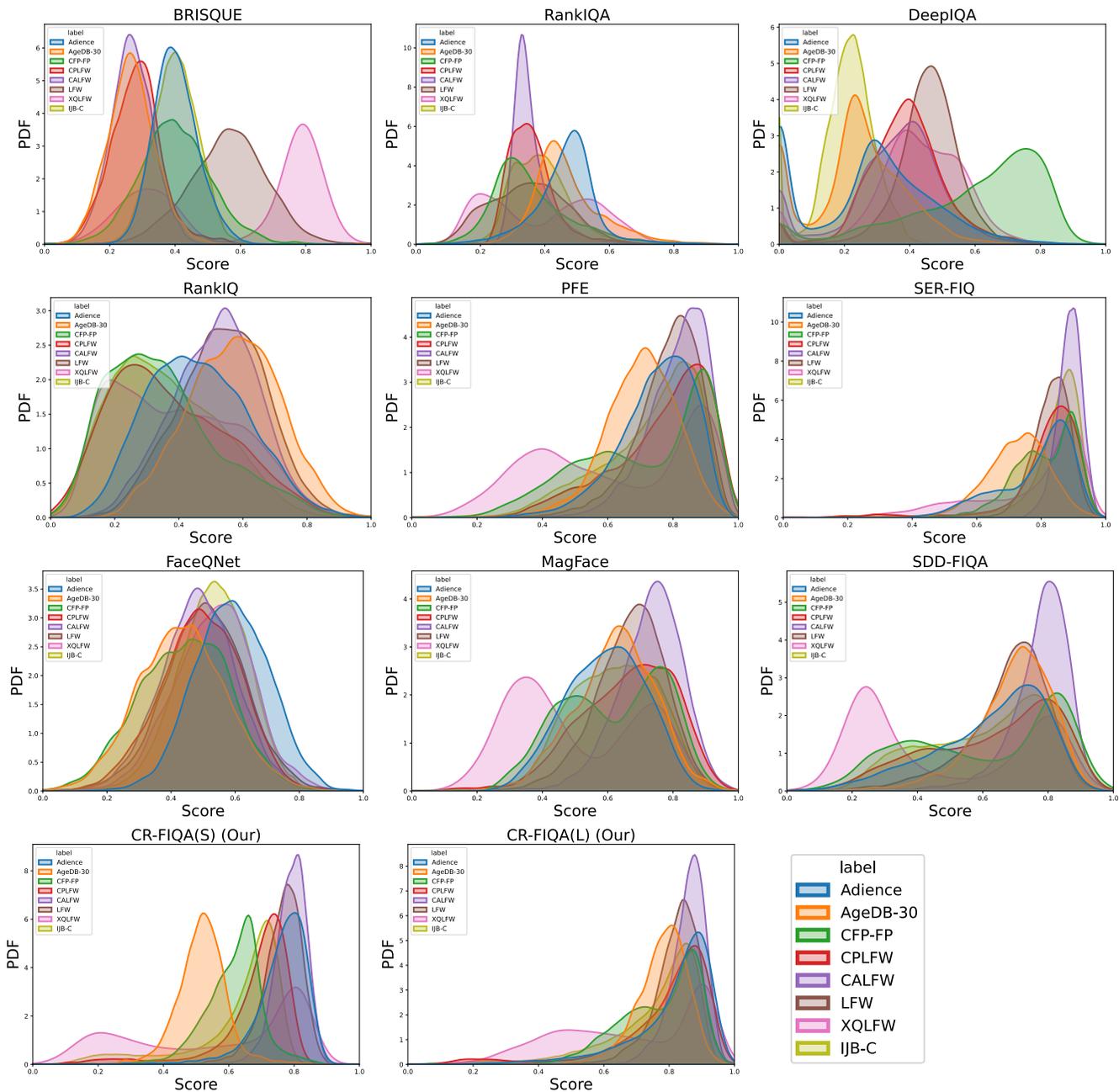

\centering
\foreach \method in {BRISQUE, RankIQA, DeepIQA}
    {\includegraphics[width=0.32\linewidth]{images/score_distribution/\method.pdf}}
\foreach \method in {RankIQ, PFE, SER-FIQ}
    {\includegraphics[width=0.32\linewidth]{images/score_distribution/\method.pdf}}
\foreach \method in {FaceQNet, MagFace, SDD-FIQA}
    {\includegraphics[width=0.32\linewidth]{images/score_distribution/\method.pdf}
    }
\foreach \method in {CR-FIQASOur, CR-FIQALOur, legend_score_dist}
    {\includegraphics[width=0.32\linewidth]{images/score_distribution/\method.pdf}
    }
\caption{Quality score distribution of the evaluation benchmarks achieved by our CR-FIQA and the SOTA methods (all normalized to have values between 0 and 1). }
\label{fig:score_distribution}
\end{figure*}

Figure \ref{fig:image_examples} shows sample images of the evaluation benchmarks with quality score values obtained from our CR-FIQA  the SOTA methods.  These images in Figure \ref{fig:image_examples} illustrate samples of different benchmarks with quality score values. It is important to mention that, although the quality scores are normalized between 0 and 1,  the higher quality score values across FIQA methods do not mean that the method points out a relative higher quality estimation than the other methods. For example, SER-FIQ method resulted always in relatively high quality score value when it is compared to other SOTA methods. However, as show in Figure \ref{fig:score_distribution}, the quality score value range of SER-FIQ is higher when compared to other SOTA methods.

\subsection{FIQA performance as ERC (FNMR at FMR1e-4 vs reject) curves}
\begin{figure*}[htb]
\centering
\foreach \db in {Adience, AgeDB-30, CFP-FP}{
    \foreach \model in {ArcFace,ElasticFace}
        {\includegraphics[scale=0.40]{images/Ercs/\db/\db_0.0001_\model.pdf}}
}
\includegraphics[scale=0.40]{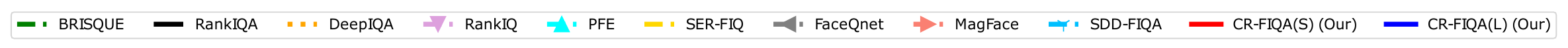}
\caption{ERC (FNMR at FMR1e-4 vs reject) curves for ArcFace and ElasticFace on Adience, AgeDB-30 and CFP-FP benchmarks. The proposed CR-FIQA(L) and  CR-FIQA(S) are marked with solid blue and red lines, respectively. CR-FIQA leads to lower verification error, when rejecting a fraction of images, of the lowest quality, in comparison to SOTA methods (faster decaying curve) under most experimental settings.}
\label{fig:erc_fmr10000_1}
\end{figure*}

\begin{figure*}[htb]
\centering
\foreach \db in {Adience, AgeDB-30, CFP-FP}{
    \foreach \model in { MagFace, CurricularFace}
        {\includegraphics[scale=0.40]{images/Ercs/\db/\db_0.0001_\model.pdf}}
}
\includegraphics[scale=0.40]{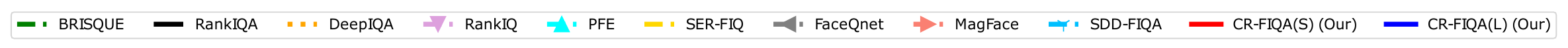}
\caption{ERC (FNMR at FMR1e-4 vs reject) curves for  MagFace and CurricularFace on Adience, AgeDB-30 and CFP-FP benchmarks. The proposed CR-FIQA(L) and  CR-FIQA(S) are marked with solid blue and red lines, respectively. CR-FIQA leads to lower verification error, when rejecting a fraction of images, of the lowest quality, in comparison to SOTA methods (faster decaying curve) under most experimental settings.}
\label{fig:erc_fmr10000_2}
\end{figure*}

\begin{figure*}[htb]
\centering
\foreach \db in {LFW, CALFW, CPLFW}{
    \foreach \model in {ArcFace,ElasticFace}
        {\includegraphics[scale=0.40]{images/Ercs/\db/\db_0.0001_\model.pdf}}
}
\includegraphics[scale=0.40]{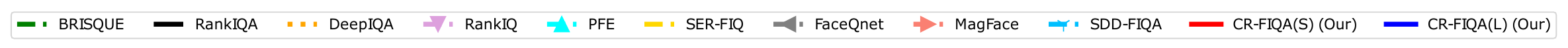}
\caption{ERC (FNMR at FMR1e-4 vs reject) curves for ArcFace and ElasticFace on LFW, CALFW and CPLFW benchmarks. The proposed CR-FIQA(L) and  CR-FIQA(S) are marked with solid blue and red lines, respectively. CR-FIQA leads to lower verification error, when rejecting a fraction of images, of the lowest quality, in comparison to SOTA methods (faster decaying curve) under most experimental settings.}
\label{fig:erc_fmr10000_3}
\end{figure*}

\begin{figure*}[htb]
\centering
\foreach \db in {LFW, CALFW, CPLFW}{
    \foreach \model in { MagFace, CurricularFace}
        {\includegraphics[scale=0.40]{images/Ercs/\db/\db_0.0001_\model.pdf}}
}
\includegraphics[scale=0.40]{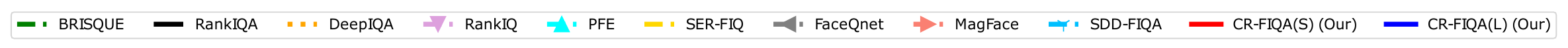}
\caption{ERC (FNMR at FMR1e-4 vs reject) curves for  MagFace and CurricularFace on LFW, CALFW and CPLFW benchmarks. The proposed CR-FIQA(L) and  CR-FIQA(S) are marked with solid blue and red lines, respectively. CR-FIQA leads to lower verification error, when rejecting a fraction of images, of the lowest quality, in comparison to SOTA methods (faster decaying curve) under most experimental settings.}
\label{fig:erc_fmr10000_4}
\end{figure*}

Figures \ref{fig:erc_fmr10000_1}, \ref{fig:erc_fmr10000_2}, \ref{fig:erc_fmr10000_3} and \ref{fig:erc_fmr10000_4} present ERC (FNMR at FMR1e-4 vs reject) curves for all the evaluation settings. These ERC curves illustrates the curves producing the AUC (FNMR at FMR1e-4) presented in Table 1 of the main submission. 
Such ERC curves are shown in Figure 4 in the main submission and Figure 18 in supplementary material on FNMR at FMR1e-3 and discussed in details in Section 6. However, we present also in this supplementary material the ERC curves on another FNMR, FNMR at FMR1e-4. These ERC curves also correspond to the AUC values presented in Table 1 of the main submission.

\subsection{FIQA performance as ERC (FNMR at FMR1e-3 vs reject) curves using MagFace and CurricularFace FR models}
Figure \ref{fig:erc_sp} presents ERC (FNMR at FMR1e-3 vs reject) curves for all the evaluation benchmarks using MagFace and CurricularFace FR models. These ERC curves also correspond to the AUC values presented in Table 1 of the main submission and discussed in details in Section 6.

\subsection{CR-FIQA as feature extraction}
The evaluation of CR-FIQA(L) backbone as feature extraction, which is not the goal of this work, on mainstream FR benchmarks is presented in Table \ref{tab:fr}. The considered benchmarks are LFW \cite{LFWTech}, AgeDB-30 \cite{agedb}, CFP-FP \cite{cfp-fp}, CALFW \cite{CALFW}, CPLFW \cite{CPLFWTech} and IJB-C \cite{ijbc}.
We followed the evaluation metrics defined in the utilized benchmarks as follows: LFW (accuracy), CALFW (accuracy), CPLFW (accuracy), CFP-FP (accuracy), AgeDB-30 (accuracy) and IJB-C (TAR at FAR1e-4).
Although,  the presented solution in this paper does not aim, and is not presented as, a solution to extract face embeddings, but rather a FIQA solution, the reported evaluation results (Table \ref{tab:fr}) are very comparable to the recent SOTA models trained under a similar training setting and only using the face recognition loss.

\subsection{Datasets}
This section presents the description and license information of the used datasets in our work.

\textbf{Adience \cite{Adience}:} Adience was used to estimate the age and gender from face images acquired in challenging and in the wild conditions. Adience dataset contains 26,580 images across 2,284 identities, where the images were captured as close to the real-world condition as possible, under all variations in appearance, pose, illuminations, and image quality. Adience license is limited to research purposes only. Detailed information on database creation and licensing can be found in \cite{Adience} and \url{https://talhassner.github.io/home/projects/Adience/Adience-main.html}.

\textbf{AgeDB-30 \cite{agedb}:} AgeDB is an in-the-wild dataset for age-invariant face verification evaluation, containing 16,488 images of 568 identities. Every image is annotated with respect to the identity, age, and gender attribute. In our case, we report the performance for AgeDB-30 (30 years age gap) as it is the most reported and challenging subset of AgeDB. More details on the collection process can be found in \cite{agedb} and the details on the license are presented in \url{https://ibug.doc.ic.ac.uk/resources/agedb/}.

\textbf{LFW \cite{LFWTech}}: Labeled Faces in the Wild (LFW) is an unconstrained face verification dataset. The LFW contains 13,233 images of 5749 identities collected from the web. The LFW is licensed under CC-BY-4.0, and more information on database creation can be found in \cite{LFWTech} and \url{http://vis-www.cs.umass.edu/lfw/}.

\textbf{CFP-FP \cite{cfp-fp}:} Celebrities in Frontal-Profile in the Wild (CFP-FP) \cite{cfp-fp} dataset addresses the comparison between frontal and profile faces. CFP-FP dataset contains 7,000 images across 500 identities, where 10 frontal and 4 profile image per identity.
More information can be found in \cite{cfp-fp} and \url{http://www.cfpw.io/}.

\textbf{CALFW \cite{CALFW}:} The Cross-age LFW (CALFW) dataset \cite{CALFW} is based on LFW with a focus on comparison pairs with the age gap, however not as large as AgeDB-30. Age gap distribution of the CALFW is provided in \cite{CALFW}. It contains 3000 genuine comparisons, and the negative pairs are selected of the same gender and race to reduce the effect of attributes. The detailed information on database creation can be found in \cite{CALFW} and \url{http://whdeng.cn/CALFW/}.

\textbf{CPLFW \cite{CPLFWTech}:} The Cross-Pose LFW (CPLFW) dataset \cite{CPLFWTech} is based on LFW with a focus on comparison pairs with pose differences. CPLFW contains 3000 genuine comparisons, while the negative pairs are selected of the same gender and race. More information can be found in \cite{CPLFWTech} and \url{http://whdeng.cn/CPLFW/}.

\textbf{XQLFW \cite{XQLFW}:} The Cross-Quality LFW (XQLFW) is derived from the LFW dataset. The XQLFW maximizes the quality difference, which contains only more realistic synthetically degraded images when necessary and is used to investigate the influence of image quality. XQLFW is licensed under the MIT License, and the detailed information can be found in \cite{XQLFW} and \url{https://martlgap.github.io/xqlfw/}.

\textbf{IJB-C \cite{ijbc}:} The IARPA Janus Benchmark–C (IJB-C) \cite{ijbc} is a video-based face recognition dataset provided by the Nation Institute for Standards and Technology (NIST). It is an extension of the IJB-B \cite{ijbb} dataset with a total of 31,334 still images and 117,542 frames of 11,779 videos across 3531 identities. IJB-C is made available under different Creative Commons license variants. Detailed information on database creation can be found in \cite{ijbc} and \url{https://www.nist.gov/programs-projects/face-challenges}.

\textbf{CASIA-WebFace \cite{casia_webface}:}
CASIA-Webface consists of 494,141 face images from 10,757 different identities. A prepossessed (aligned and cropped) version of CASIA-WebFace is available in InsightFace (\url{https://insightface.ai/}) repository  under Dataset-Zoo
\url{https://github.com/deepinsight/insightface/tree/master/recognition/_datasets_ }. 
The code and the databases of InsightFace is under MIT licence (\url{https://github.com/deepinsight/insightface/blob/master/LICENSE}).

\textbf{MS1MV2 \cite{guo2016ms,deng2019arcface}:} The MS1MV2 is a refined version \cite{deng2019arcface} of the MS-Celeb-1M \cite{guo2016ms} containing 5.8M images of 85K identities. A prepossessed (aligned and cropped) version of MS1MV2 is available in InsightFace (\url{https://insightface.ai/} repository under Dataset-Zoo
\url{https://github.com/deepinsight/insightface/tree/master/recognition/_datasets_ }.
The code and the database of InsightFace is under MIT licence (\url{https://github.com/deepinsight/insightface/blob/master/LICENSE}).

\subsection{Use of existing assets}
The results of the SOTA FIQA methods are produced based on the official code provided by each of these works. Table \ref{tab:assets} presents the used SOTA methods along with link to their code repositories and licences. 

The utilized FR models  to report the verification performance at different quality rejection rates are  ArcFace \cite{deng2019arcface}, ElasticFace (ElasticFace-Arc) \cite{elasticface}, MagFace \cite{MagFace}, and CurricularFace \cite{curricularFace}.  The link to the official code repository and license for each of the employed FR models are provided in the following: 
\begin{itemize}
    \item ArcFace \cite{deng2019arcface} is provided under MIT license \url{https://github.com/deepinsight/insightface/blob/master/LICENSE} and the official pretrained model and code is published under the link \url{https://github.com/deepinsight/insightface}.
    \item MagFace \cite{MagFace} is provided under Apache License 2.0 \url{https://github.com/IrvingMeng/MagFace/blob/main/LICENSE} and the official pretrained model and code is published under the link \url{https://github.com/IrvingMeng/MagFace}. 
    \item CurricularFace \cite{curricularFace} is provided underMIT license \url{https://github.com/HuangYG123/CurricularFace/blob/master/LICENSE} and the official pretrained model and code is published under the link \url{https://github.com/HuangYG123/CurricularFace/}.
    \item ElasticFace \cite{elasticface} is provided under Attribution-NonCommercial-ShareAlike 4.0 
International (CC BY-NC-SA 4.0) license \url{https://github.com/fdbtrs/ElasticFace/blob/main/README.md} and the official pretrained model and code is published under the link \url{lhttps://github.com/fdbtrs/ElasticFace}.
\end{itemize}

\begin{table*}[]
\centering
\resizebox{\textwidth}{!}{%
\begin{tabular}{|l|l|l|l|l|l|l|}
\hline
\multirow{2}{*}{Model} &
  \multirow{2}{*}{\begin{tabular}[c]{@{}l@{}}LFW\\ Acc (\%)\end{tabular}} &
  \multirow{2}{*}{\begin{tabular}[c]{@{}l@{}}AgeDB-30\\ Acc (\%)\end{tabular}} &
  \multirow{2}{*}{\begin{tabular}[c]{@{}l@{}}CFP-FP\\ Acc (\%)\end{tabular}} &
  \multirow{2}{*}{\begin{tabular}[c]{@{}l@{}}CALFW\\ Acc (\%)\end{tabular}} &
  \multirow{2}{*}{\begin{tabular}[c]{@{}l@{}}CPLFW\\ Acc (\%)\end{tabular}} &
  \multirow{2}{*}{\begin{tabular}[c]{@{}l@{}}IJB-C\\ TAR at FAR\=1e-4\end{tabular}} \\
                   &       &       &       &       &       &       \\ \hline
ArcFace \cite{deng2019arcface}           & 99.82 & 98.15 & 98.27 & 95.45 & 92.08 & 96.28 \\ \hline
ElasticFace \cite{elasticface}        & 99.80 & 98.35 & 98.67 & 96.17 & 93.27 & 96.49 \\ \hline
MagFace     \cite{MagFace}       & 99.83 & 98.17 & 98.46 & 96.15 & 92.87 & 96.65 \\ \hline
CurricularFace  \cite{curricularFace}   & 99.80 & 98.32 & 98.37 & 96.20 & 93.13 & 96.58 \\ \hline
CR-FIQA (L) (Ours) & 99.80 & 98.17 & 98.49 & 96.15 & 92.90 & 96.23 \\ \hline
\end{tabular}%
}
\caption{The verification performances of CR-FIQA (L) as feature extraction models on mainstream bookmarks and compared to the recent SOTA face recognition models. }
\label{tab:fr}
\end{table*}

\begin{table*}[]
\centering
\resizebox{\textwidth}{!}{%
\begin{tabular}{|l|l|l|}
\hline
Method &
  Code link &
  License \\ \hline
SER-FIQA \cite{SERFIQ} &
  \url{https://github.com/pterhoer/FaceImageQuality} &
  \begin{tabular}[c]{@{}l@{}}Attribution-NonCommercial-ShareAlike 4.0 International (CC BY-NC-SA 4.0) license\\  \url{https://github.com/pterhoer/FaceImageQuality/blob/master/README.md}\end{tabular} \\ \hline
FaceQnet \cite{faceqnetv1} &
  \url{https://github.com/uam-biometrics/FaceQnet} &
  no specific license provided by the authors \\ \hline
MagFace \cite{MagFace} &
  \url{https://github.com/IrvingMeng/MagFace} &
  \begin{tabular}[c]{@{}l@{}}Apache License 2.0 \\ \url{https://github.com/IrvingMeng/MagFace/blob/main/LICENSE}\end{tabular} \\ \hline
SDD-FIQA \cite{SDDFIQA}&
  \url{https://github.com/Tencent/TFace/tree/quality} &
  \begin{tabular}[c]{@{}l@{}}Extension of Apache License Version 2.0\\ \url{https://github.com/Tencent/TFace/blob/master/License.txt}\end{tabular} \\ \hline
rankIQ \cite{RANKIQ_FIQA} &
  \url{https://jschenthu.weebly.com/projects.html} &
  \begin{tabular}[c]{@{}l@{}}This toolbox is made available for research purpose only as stated\\  in README.md of code webpage\end{tabular} \\ \hline
BRISQUE \cite{BRISQE_IQA} &
 \url{ http://live.ece.utexas.edu/research/quality/BRISQUE_release.zip} &
  Free usage is stated in the readme file contained in the project \\ \hline
PFE  \cite{PFE_FIQA}&
  \url{https://github.com/seasonSH/Probabilistic-Face-Embeddings} &
  \begin{tabular}[c]{@{}l@{}}MIT License \\ \url{https://github.com/dmaniry/deepIQA/blob/master/LICENSE}\end{tabular} \\ \hline
rankIQA  \cite{liu2017rankiqa} &
  \url{https://github.com/xialeiliu/RankIQA} &
  \begin{tabular}[c]{@{}l@{}}MIT License\\ \url{https://github.com/xialeiliu/RankIQA/blob/master/LICENSE} \end{tabular} \\ \hline
DeepIQA  \cite{DEEPIQ_IQA} &
  \url{https://github.com/dmaniry/deepIQA} &
  \begin{tabular}[c]{@{}l@{}}MIT License\\ \url{https://github.com/dmaniry/deepIQA/blob/master/LICENSE} \end{tabular} \\ \hline
\end{tabular}%
}
\caption{ The official released code links and licenses of the FIQA methods reported in this work. The results of the  FIQA methods in the main submission are produced and reported based on their official released code and strictly following their licenses.}
\label{tab:assets}
\end{table*}

\subsection{Release of implementation and pre-trained models}

The implementation and pre-trained models will be released publicly under the Attribution-NonCommercial-ShareAlike 4.0 
International (CC BY-NC-SA 4.0) license. 
A copy of the code is under \url{https://github.com/fdbtrs/CR-FIQA}

\subsection{Potential societal impacts}
We stress that our efforts in the advancement of FIQA and thus, face recognition, are aimed at enhancing the security, convenience, and life quality of the members of society,  e.g. enabling convenient access to financial and health services \cite{eadhaar} and enhancing the security of border checks within clear legal frameworks and users consent \cite{GDPR_practice,bipa}. We acknowledge, however reject, the possible malicious or illegal use of this and other machine learning-based technologies. 
Such a use of face recognition can involve the processing of face images for barometric recognition purposes out of legal framework and without the consent of the individual to create user/group profiles or the not consent use of face recognition in functionalities beyond the identity recognition itself \cite{DBLP:journals/tifs/MedenRTDKSRPS21}.

\subsection{Limitation of the proposed approach}

Unlike methods where the FIQA does not require to train a quality regression \cite{SERFIQ,MagFace,PFE_FIQA} our CR-FIQA requires a training a regression. However, this only required to be done once and the resulting model can be used to estimate quality for multiple efficiently FR models as demonstrated by the result.

\begin{figure*}[htb]
\centering
\foreach \db in { XQLFW, IJB-C}{
    \foreach \model in {ArcFace,ElasticFace}
        {\includegraphics[scale=0.40]{images/Ercs/\db/\db_0.0001_\model.pdf}}
}
\includegraphics[scale=0.40]{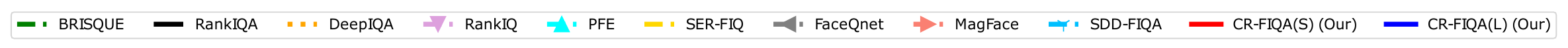}
\caption{ERC (FNMR at FMR1e-4 vs reject) curves for ArcFace and ElasticFace on XQLFW and IJB-C benchmarks. The proposed CR-FIQA(L) and  CR-FIQA(S) are marked with solid blue and red lines, respectively. CR-FIQA leads to lower verification error, when rejecting a fraction of images, of the lowest quality, in comparison to SOTA methods (faster decaying curve) under most experimental settings.}
\label{fig:erc3_fmr10000}
\end{figure*}

\begin{figure*}[htb]
\centering
\foreach \db in { XQLFW, IJB-C}{
    \foreach \model in { MagFace, CurricularFace}
        {\includegraphics[scale=0.40]{images/Ercs/\db/\db_0.0001_\model.pdf}}
}
\includegraphics[scale=0.40]{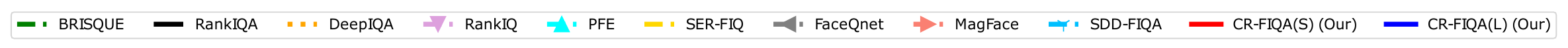}
\caption{ERC (FNMR at FMR1e-4 vs reject) curves for MagFace and CurricularFace on XQLFW and IJB-C benchmarks. The proposed CR-FIQA(L) and  CR-FIQA(S) are marked with solid blue and red lines, respectively. CR-FIQA leads to lower verification error, when rejecting a fraction of images, of the lowest quality, in comparison to SOTA methods (faster decaying curve) under most experimental settings.}
\label{fig:erc3_1_fmr10000}
\end{figure*}

\begin{figure*}[h!]
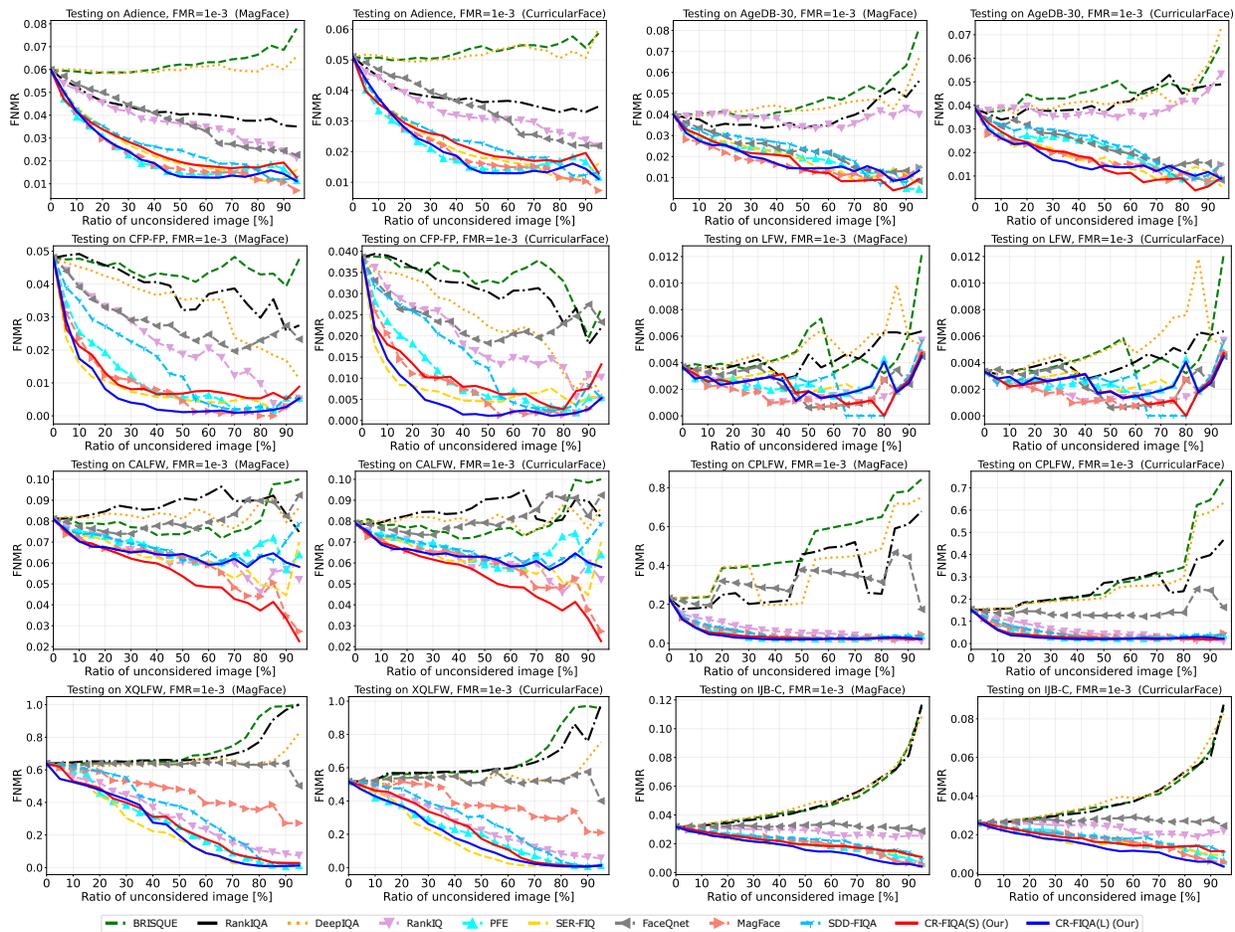

\centering
\foreach \db in {Adience, AgeDB-30, CFP-FP, LFW, CALFW, CPLFW, XQLFW, IJB-C}{
    \foreach \model in {MagFace, CurricularFace}
        {\includegraphics[scale=0.20]{images/Ercs/\db/\db_0.001_\model.pdf}}
}
\includegraphics[scale=0.35]{images/Ercs/erc_legend.pdf}
\caption{ERC (FNMR at FMR1e-3 vs reject) curves for all evaluated benchmarks using MagFace and CurricularFace FR models corresponding to Table 1 and complementary to the ERC curves in Figure 4 in main submission results.
The proposed CR-FIQA(L) and  CR-FIQA(S) are marked with solid blue and red lines, respectively. 
CR-FIQA leads to lower verification error, when rejecting a fraction of images, of the lowest quality, in comparison to SOTA methods (faster decaying curve) under most experimental settings.
}
\label{fig:erc_sp}
\end{figure*}

\end{document}